\documentclass[10pt,twocolumn,letterpaper]{article}

\pdfoutput=1
\usepackage{iccv}              

\usepackage[ruled,vlined,linesnumbered]{algorithm2e}
\usepackage{algpseudocode}

\usepackage{epsfig}
\usepackage{caption}
\usepackage{subcaption}
\usepackage{array}
\usepackage{xcolor}
\usepackage{multirow}
\usepackage{multicol}
\usepackage{ragged2e}
\usepackage{arydshln}
\usepackage{indentfirst}
\usepackage{graphicx}
\usepackage{footmisc}
\usepackage{soul}
\usepackage[normalem]{ulem}
\usepackage[accsupp]{axessibility}  

\makeatletter
\newcommand*\bigcdot{\mathpalette\bigcdot@{1.2}}
\newcommand*\bigcdot@[2]{\mathbin{\vcenter{\hbox{\scalebox{#2}{$\m@th#1\bullet$}}}}}
\makeatother

\definecolor{LAA-Net}{HTML}{f27452}
\definecolor{MultiAtt.}{HTML}{4c4cff}
\definecolor{SBI}{HTML}{64ae64}
\definecolor{Xception}{HTML}{a64ca6}
\definecolor{RCCE}{HTML}{a6a6a6}
\definecolor{CADDM}{HTML}{fbde6b}

\definecolor{darkgreen}{rgb}{0,0.5,0} 
\definecolor{purple}{rgb}{1,0,1} 

\newcolumntype{H}{>{\setbox0=\hbox\bgroup}c<{\egroup}@{}}

\definecolor{iccvblue}{rgb}{0.21,0.49,0.74}
\usepackage[pagebackref,breaklinks,colorlinks,allcolors=iccvblue]{hyperref}

\def\paperID{14341} 
\def\confName{ICCV}
\def\confYear{2025}

\title{Vulnerability-Aware Spatio-Temporal Learning for Generalizable Deepfake Video Detection}

\author{Dat NGUYEN$^{\triangleleft}$, Marcella ASTRID$^{\triangleleft}$, Anis KACEM$^{\triangleleft}$, Enjie GHORBEL$^{\triangleleft,\rtimes}$, Djamila AOUADA$^{\triangleleft}$ \\
CVI$^2$, SnT, University of Luxembourg$^{\triangleleft}$ \\
Cristal Laboratory, National School of Computer Sciences, University of Manouba$^{\rtimes}$ \\
{\tt\small \{dat.nguyen,marcella.astrid,anis.kacem,djamila.aouada\}@uni.lu enjie.ghorbel@isamm.uma.tn}
}

\begin{document}
\maketitle
\begin{abstract}

Detecting deepfake videos is highly challenging given the complexity of characterizing spatio-temporal artifacts. Most existing methods rely on binary classifiers trained using real and fake image sequences, therefore hindering their generalization capabilities to unseen generation methods. Moreover, with the constant progress in generative Artificial Intelligence (AI), deepfake artifacts are becoming imperceptible at both the spatial and the temporal levels, making them extremely difficult to capture.  To address these issues, we propose a fine-grained deepfake video detection approach called FakeSTormer that enforces the modeling of subtle spatio-temporal inconsistencies while avoiding overfitting. Specifically, we introduce a multi-task learning framework that incorporates two auxiliary branches for explicitly attending artifact-prone spatial and temporal regions. Additionally, we propose a video-level data synthesis strategy that generates pseudo-fake videos with subtle spatio-temporal artifacts, providing high-quality samples and hand-free annotations for our additional branches. Extensive experiments on several challenging benchmarks demonstrate the superiority of our approach compared to recent state-of-the-art methods.
The code is available at \url{https://github.com/10Ring/FakeSTormer}.
\end{abstract}
    
\section{Introduction}
\label{sec:intro}

With the advances in generative modeling~\cite{gan, fsganv2}, deepfake videos have become alarmingly realistic. Despite their interest in several applications, such as entertainment and education, this type of technology also raises societal concerns~\cite{russia-ukraine-war, korea_deepfake_porn, 25m_video_call}. There is therefore an urgency for developing effective deepfake detection methods. 

In the literature, several deepfake detection techniques aim to model spatial artifacts by treating each frame independently~\cite{ff++, fxray, multi-attentional, sladd, ete_recons, sbi, caddm, ost, nguyen2024fakeformer, laa_net, untag}. While this is reasonable when dealing with frame-level generation methods~\cite{face2face, fake_f_gen1, fake_f_gen3}, it becomes less adequate in the presence of video-level manipulation techniques~\cite{f_v_gen1, f_v_gen2, f_v_gen3}, where temporal and spatial artifacts are \textit{intertwined}. 

\begin{figure}
    \centering
    \includegraphics[width=0.83\linewidth]{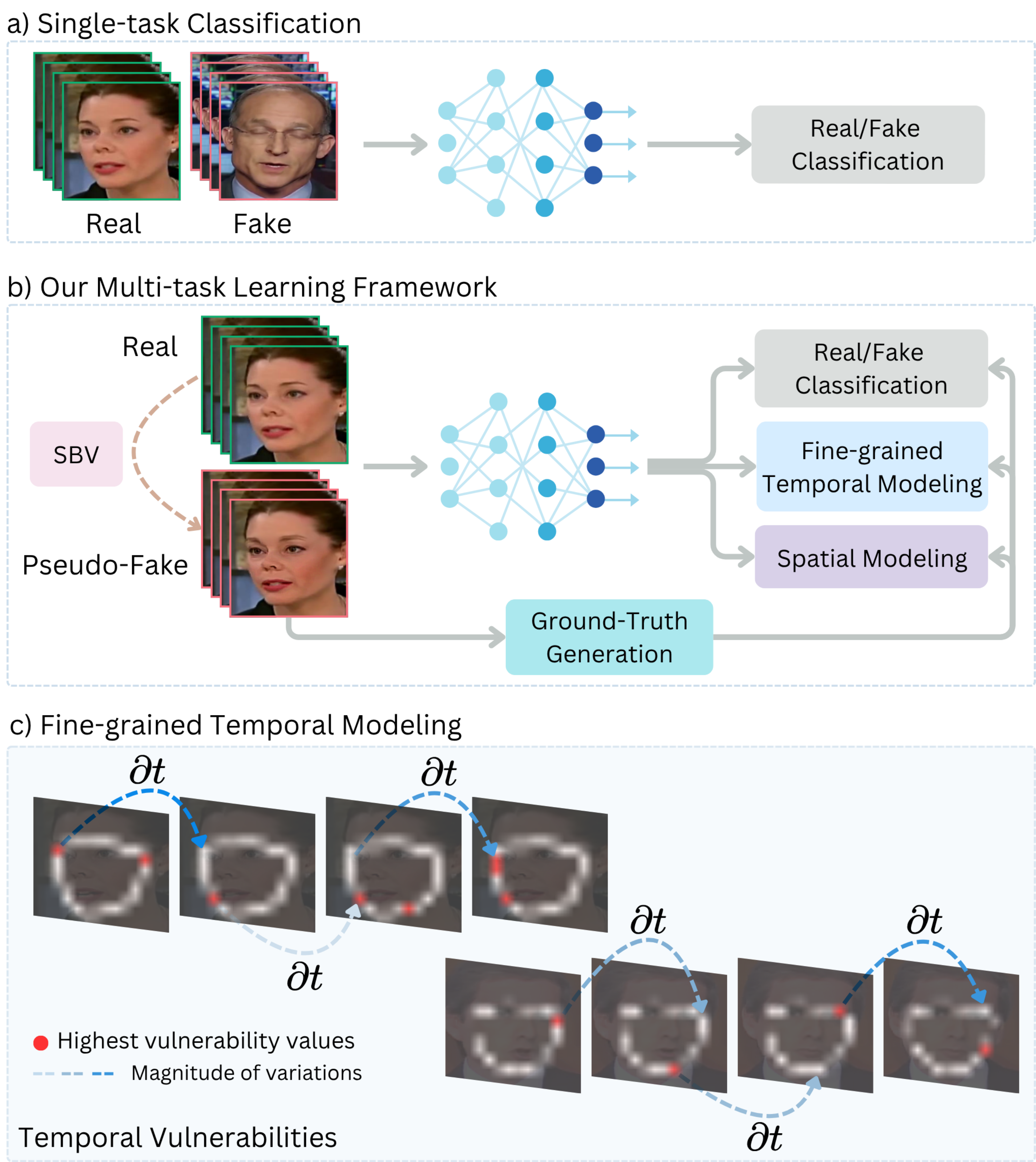}
    \vspace{-1mm}
    \caption{a) Traditional video-based methods~\cite{ftcn, altfreezing, lipforensics, tall_swin, istvt, sp_temp_features, ST_dropout, stil} versus b) the proposed multi-task learning framework; c) Visualization of the temporal vulnerabilities. Note that only some temporal locations are shown.}
    \label{fig:motivation}
    \vspace{-3mm}
\end{figure}

For that reason, researchers have explored video-based deepfake detection methods capable of modeling spatio-temporal artifacts~\cite{stil, sp_temp_features, ftcn, istvt, lipforensics, ST_dropout, altfreezing}. Those methods mainly rely on a deep neural network formed by a single binary classification branch that is trained using a fixed dataset with real and fake data (see Figure~\ref{fig:motivation}-a). As a result, they suffer from two main limitations, namely: (1) \textbf{The lack of generalizability -} As highlighted in~\cite{untag, laa_net, fxray, sbi}, models trained with a standalone binary classifier tend to overfit the type of deepfakes they are trained on, resulting in poor generalization to unseen manipulations; (2) \textbf{The lack of robustness to high-quality (HQ) deepfake videos -} The quality of deepfake videos is improving continuously, resulting in subtle spatio-temporal artifacts. 
As such, vanilla single-branch architectures trained solely with binary supervision fail to fully capture them, necessitating the design of appropriate attention mechanisms.


To address the generalization issue, video-level data synthesis approaches~\cite{STC, stsbv, plug_play} have been introduced to encourage models to learn more generic representations. However, these methods usually simulate exaggerated temporal variations that are inherently different from artifacts in hyper-realistic deepfake videos. 
On the other hand, to model localized spatio-temporal inconsistencies, some recent methods~\cite{istvt, LFGDIN} have used dedicated architectures integrating an implicit attention mechanism. Nevertheless, these models still rely solely on a binary classifier with no guarantee of extracting artifact-prone \textit{fine-grained} traces.


Interestingly, in image-based deepfake detection, it has been recently demonstrated that the use of a tailored multi-task learning framework for explicitly attending artifact-prone small regions coupled with a subtle data synthesis strategy can be a way to enhance generalization and, at the same time, robustness to high-quality deepfakes ~\cite{ laa_net,nguyen2024fakeformer}. Nevertheless, such an approach has been disregarded in the field of video-level deepfake detection, as its extension to the video level is not straightforward. In particular, it would necessitate the characterization of subtle temporal artifacts that are inherently different from spatial ones, within both the multi-task learning framework and the data synthesis.

In this paper, we redefine deepfake video detection as a fine-grained detection task by proposing a multi-branch network that leverages synthesized data and incorporates specialized learning objectives specifically targeting both subtle spatial and temporal artifacts.
As shown in Figure~\ref{fig:motivation}-b, a novel multi-task learning framework, termed FakeSTormer is introduced. It is formed by two auxiliary parallel branches in addition to the standard classification head, namely: \textbf{(1) a regression temporal branch} incorporating an explicit attention that aims at locating the vulnerability-prone temporal locations. 
It has been shown that regressing spatial vulnerabilities in specific points~\cite{laa_net} or patches~\cite{nguyen2024fakeformer} can help improve the generalizability of a deepfake detector model. We refer to the definitions given in~\cite{nguyen2024fakeformer,laa_net} which describe: \textit{``vulnerable patches/points as the patches/points that are the most likely to embed blending artifacts"}.
To generalize this concept to the temporal domain, we propose locating temporal high changes in spatial vulnerable patches (see Figure~\ref{fig:motivation}-c). 
\textbf{(2) a spatial branch} to ensure a balance between the spatial and the temporal domains. In fact, detecting spatial artifacts in addition to temporal ones is crucial~\cite{altfreezing, istvt}. For that purpose, we propose predicting frame-wise spatial vulnerabilities.

To create hand-free ground truths for the proposed branches, we introduce a HQ video-level data synthesis algorithm, called ``Self-Blended Video (SBV)'', inspired by ``Self-Blended Image (SBI)~\cite{sbi}'', enforcing temporal coherence using two proposed modules on top of SBI (detailed in Section.~\ref{subsec:data}). Our experiments demonstrate that simply training a baseline classification model on SBV enables achieving on par performance \wrt state-of-the-art (SOTA), highlighting the effectiveness of SBV. Finally, for enhancing spatial and temporal modeling, we revisit the TimeSformer~\cite{timesformer} architecture that we use as our backbone. 
In particular, we leverage TimeSformer’s decomposed temporal and spatial attention on embedded patches, appending classification tokens for each frame and for each patch across frames, rather than a single token for the entire video. These classification tokens are then used within the spatial and classification heads, while the embedded patches are used within the temporal head. 
Extensive experiments on several well-known deepfake detection benchmarks show that our method outperforms the existing SOTA approaches.

\vspace{0.5mm}
\noindent\textbf{Contributions.} In summary, we propose in this paper: 

\begin{itemize}
  \item A novel multi-task learning framework using only real data for fine-grained video-based deepfake detection.
  \item Two auxiliary branches that capture both temporal and spatial vulnerabilities, that are fined-grained by definition.
   \item A video-level data synthesis technique called SBV that generates high-quality pseudo-fakes and is supported by a vulnerability-driven cutout augmentation strategy to avoid overfitting specific artifact-prone regions.
    \item A revisited version of the TimeSformer~\cite{timesformer}, specifically tailored for the proposed video-based deepfake detector.
    \item Extensive experiments and analyses conducted on several challenging datasets.
    \end{itemize}

\vspace{0.5mm}
\noindent\textbf{Paper organization.} 
Section~\ref{sec:relatedW} reviews related work on video deepfake detection. Section~\ref{sec:method} describes the proposed FakeSTormer method. Section~\ref{sec:exp} presents experiments and results. Finally, Section~\ref{sec:conclu} concludes with future work.

\section{Related Work}
\label{sec:relatedW}

\noindent\textbf{Video-based Deepfake Detection.}\label{sec:deepfake_video}
As highlighted in~\cite{ftcn, altfreezing}, using a naive spatio-temporal binary classification model for video-level deepfake detection can lead the model to overfit obvious artifacts, resulting in poor generalization to unseen manipulations. To address this, FTCN~\cite{ftcn} proposes a fully temporal convolution network by reducing the spatial kernel size to one, hence decreasing the likelihood of focusing only on spatial artifacts. LipForensics~\cite{lipforensics} considers solely the mouth region,
while spatio-temporal dropout~\cite{ST_dropout} randomly removes parts of the input frames in both spatial and temporal domains.
AltFreezing~\cite{altfreezing} separates convolution layers into spatial and temporal ones, failing to model long-term dependencies. 
Instead of using convolution layers, ISTVT~\cite{istvt} utilizes a video-based Vision Transformer~\cite{ViT} with self-attention to extract longer-range correlations. 
Meanwhile, \cite{stylelatent} decomposes features into spatial and temporal components. TALL~\cite{tall_swin} employs an image-level deepfake detector by converting video frames into a thumbnail layout. Despite being promising, most of the aforementioned methods solely rely on a single binary classifier that implicitly guides the feature extraction. As highlighted in the literature on image-based deepfake detection~\cite{fxray, sladd, laa_net, caddm}, this approach might lead to overfitting specific artifacts present in training datasets. Moreover, the absence of an explicit attention mechanism to spatio-temporal artifact-prone regions can lead to poor robustness to high-quality artifacts.

\vspace{0.8mm}
\noindent\textbf{Data Synthesis.}
A highly effective approach for enhancing the generalizability of deepfake detectors is training models with synthesized data. While frame-level solutions have been extensively studied~\cite{sbi, fxray, cstency_learning, sladd, caddm}, video-level augmentations remains relatively underexplored. In recent works, STC~\cite{STC} generates pseudo-fake samples via time-shuffling, 
VB~\cite{plug_play} perturbs landmarks per frame without imposing temporal coherence,
while ST-SBV~\cite{stsbv} injects temporal artifacts through random face scaling and blurring over time. However, these methods often introduce exaggerated temporal distortions that differ from HQ deepfakes typically exhibiting finer temporal inconsistencies. 



\begin{figure*}
    \centering
    \includegraphics[width=0.86\linewidth]{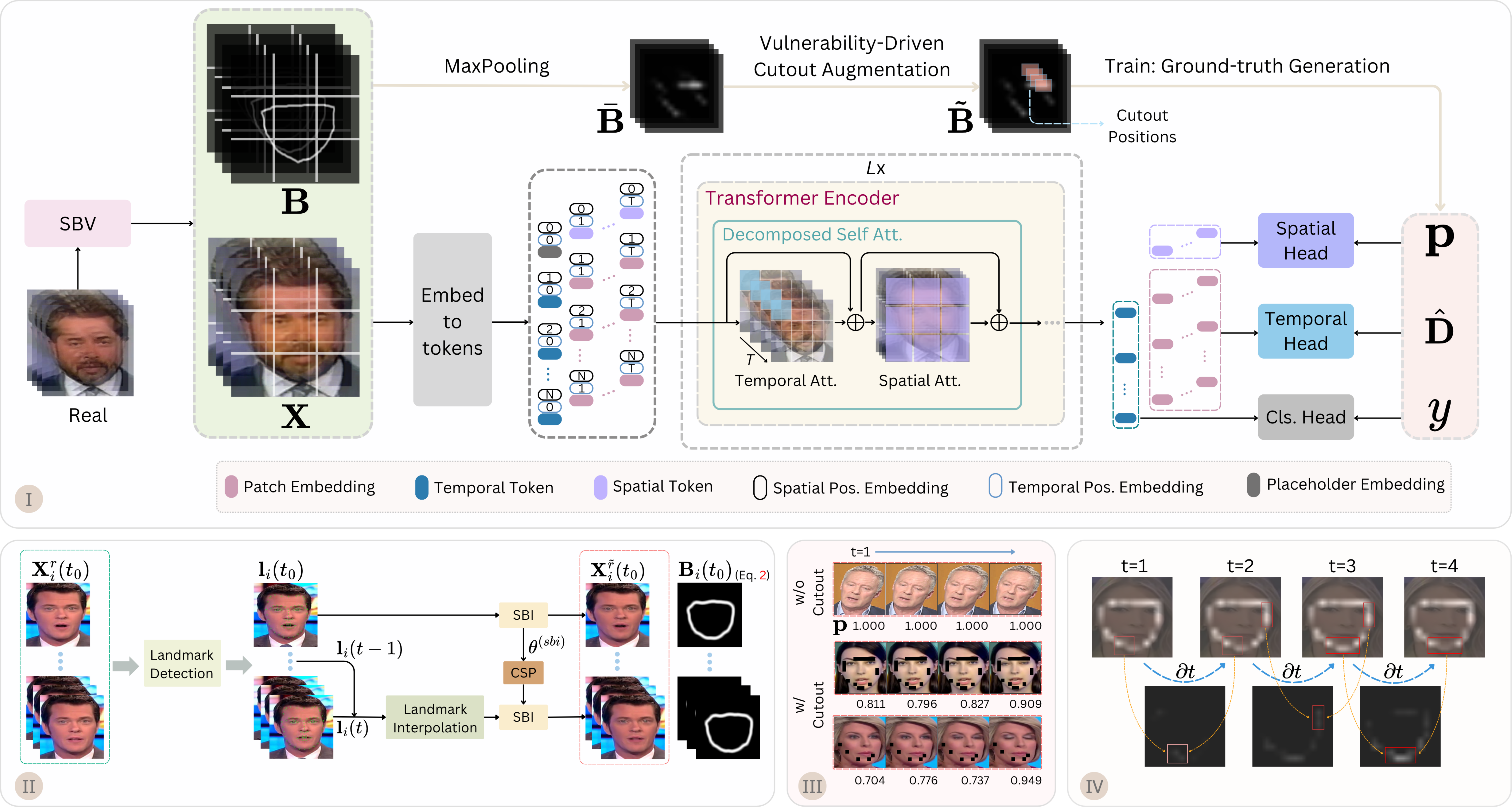}
    \vspace{-1.5mm}
    \caption{\textbf{I) Overview of the proposed framework:} Our multi-task learning framework, FakeSTormer, consists of three branches, i.e., the temporal branch ($h$), the spatial branch ($g$), and the standard classification branch ($f$). Those branches are specially designed to facilitate the disentanglement learning of spatial-temporal features. The hand-free ground-truth data to train the framework are generated based on our proposed video-level data synthesis algorithm coupled with a vulnerability-driven Cutout strategy.
    \textbf{II) Overview of generating a self-blended video:} It contains two main components, including a landmark interpolation module (\textit{LI}) and the consistent utilization of synthesized parameters (\textit{CSP}).
    \textbf{III) Examples of pseudo-fake videos:} with(\textit{w/}) and without(\textit{w/o}) vulnerability-driven Cutout and their corresponding soft labels. We apply the Cutout data augmentation at the same spatial locations throughout video frames.
    \textbf{IV) Extraction of temporal vulnerabilities:} We compute derivatives of the spatial vulnerabilities over time.
    }
    \vspace{-4mm}
    \label{fig:overview_fw}
\end{figure*}

\section{Methodology}
\label{sec:method}

Let $\mathcal V \triangleq \cup_{i=1}^{N}\{(\mathbf X_i,  y_i)\}$ be a training dataset formed by $N$ videos, where $\mathbf X_i$ denotes the $i^{\text{th}}$ video sample and $y_i$ its associated label indicating whether the clip is real ($y_i=0$) or fake  ($y_i=1$). Traditional methods~\cite{altfreezing,stil, tall_swin, lipforensics, ftcn, istvt, ST_dropout, sp_temp_features} aim to learn jointly a feature extractor $\Phi: \mathcal V \mapsto \mathcal F$ and a binary classifier $f:\mathcal F \rightarrow \{0,1\}$ by minimizing the standard binary cross-entropy (BCE) loss $\mathcal L_{BCE}(f(\Phi(\mathbf X_i)), y_i)$ using the entire training set $\mathcal V$, with $\mathcal F$ being the learned feature space. As previously discussed in~\cite{fxray, laa_net, untag, sbi} and also highlighted in Section~\ref{sec:intro}, such a strategy might lead to poor generalization capabilities to unseen generation methods while providing only binary outputs that are not interpretable.


To tackle these issues, inspired by the literature on image-level deepfake detection~\cite{fxray, laa_net, sladd, cstency_learning, caddm}, we introduce a novel multi-task learning framework called FakeSTormer that only relies on the real data subset denoted as $\mathcal V^r \subset \mathcal V$. Specifically, in addition to the binary classifier $f$, our framework includes two additional branches $h:\mathcal F \rightarrow \mathcal H$ and $g:\mathcal F \rightarrow \mathcal G$ that aim at triggering the learning of localized temporal and spatial artifact-prone features, respectively, through relevant auxiliary tasks. Note that $\mathcal H$ and $\mathcal G$ denote respectively the output spaces of $h$ and $g$. The proposed branches are depicted further in Section~\ref{sec:fakestormer}. To provide ground truth to those branches and at the same time avoid overfitting to specific manipulations, we apply to each video belonging to  $\mathcal V^r$  a data synthesis method described in Section~\ref{subsec:data}, resulting in a pseudo-fake subset denoted as $\mathcal V^{\Tilde r} $. Hence, our framework is trained using $\mathcal{\Tilde{V}} \triangleq \{\mathcal V^r \cup \mathcal V^{\Tilde r} \}$. 

\subsection{Video-Level Data Synthesis and Augmentation}
\label{subsec:data}



\paragraph{Self-Blended Video.} Blending-based data synthesis methods have demonstrated great performance in image-based deepfake detection~\cite{sbi, fxray, laa_net, sladd, ost, cstency_learning, nguyen2024fakeformer}. In fact, as the blending step is common to different manipulation types, they contribute to the improvement of the generalization aspect in deepfake detection~\cite{laa_net, fxray}. Nevertheless, such an approach has been overlooked in the context of video-based deepfake detection. Hence, we propose to extend blending-based data synthesis to the video level.
In particular, we revisit Self-Blended Image (SBI)~\cite{sbi}, given its ability to produce high-quality pseudo-fake images. 
The proposed data synthesis approach, termed Self-Blended Video (SBV), is constituted
of two main components building on top of SBI, \textit{i.e.}, a Consistent Synthesized Parameters (CSP) module followed by a Landmark Interpolation module (LI) for preserving the temporal coherence of synthesized videos, which is essential for producing high-quality synthesized videos.

Specifically, given a real video $\mathbf X_i^r \in \mathcal V^r$ formed by $T$ consecutive frames, we start by extracting a set of 2D landmarks $\mathbf L_{i} (t) = \{\mathbf l_{ij}(t)\}_{1 \leq j \leq n }$ at each instant $t$ from $\mathbf X_i^r(t)$, where $n$ refers to the number of landmarks and $ \{\mathbf l_{ij}(t)\} \in \mathbb R^{2}$.  Then, we apply SBI to the $1^{\text{st}}$ video frame denoted as $\mathbf X^r_{i}(t_0)$ to obtain a pseudo-fake image, i.e., $\mathbf X_{i}^{\Tilde{r}}(t_0)$ and a blending mask $\mathbf M_{i}(t_0)$. All related blending parameters $\mathbf{\theta}^{(sbi)}$ (e.g., ConvexHull type, Mask deformation kernels, blending ratio, etc.) are then conserved for synthesizing the remaining video frames. However, using those parameters solely cannot guarantee the temporal consistency of pseudo-fake videos since the geometry of landmarks can significantly vary over time. To mitigate the issue, we propose to re-interpolate each landmark $\mathbf l_{i}(t)$ based on $\mathbf l_{i}(t-1)$ for $t>t_0$ as follows,
\begin{equation}
\mathbf l_{i} (t) = 
\begin{cases}
\begin{split}
    \mathbf l_{i} (t-1) + \frac{\mathbf l_{i}(t) - \mathbf l_{i} (t-1)}{\operatorname{round}( d/\bar{d} )} \ , 
    \text{if } d > \tau \\
    \text{where} \hspace{1mm} d = \ \| \mathbf l_{i} (t) - \mathbf l_{i}(t-1) \|_2/n
\end{split} \\
\mathbf l_{i}(t) \text{, otherwise,} \, 
\end{cases}
\label{eq:lmk_interpolation}
\end{equation}
where $d$ represents the normalized distance between the position of a given landmark at the instants $t$ and $t-1$,
$\tau$ is a constant threshold for determining when to interpolate (interpolation intervenes only in the presence of drastic changes), 
and $\bar d$ is empirically chosen.
Overall, a higher $d$ value will push the updated point to move closer to the previous landmark position $\mathbf l_{i} (t-1)$, hence contributing to smooth the landmark position over time. However, excessive smoothing can be disadvantageous, as it can discard temporal artifacts. To address this, we use a $\operatorname{round}$ operator to incorporate slight errors. Hence, the proposed SBV data synthesis produces high-quality pseudo-fake videos incorporating subtle temporal artifacts.


As a result, we obtain the pseudo-fake video $\mathbf X_i^{\Tilde r} \in \mathcal V^{\Tilde r}$ and its blending mask sequence $\mathbf M_i \in \mathbb R^{T \times H \times W}$, with $H$ and $W$ being the image height and width, respectively. An illustration of SBV is given in Figure~\ref{fig:overview_fw}-II. Additional samples, as well as the detailed algorithm, are provided in supplementary materials. It is important to note that, despite being simple, SBV is generic and applicable to any existing video-level deepfake detection approach.

\vspace{0.5mm}
\noindent\textbf{Vulnerability-Driven Cutout Augmentation.}
\label{subsec:cutout_augmentation}
Previous works~\cite{random_erasing, rfm} have demonstrated that deep learning methods are often impacted by overfitting. Deepfake detectors might even be more sensitive to this phenomenon as deepfakes are typically characterized by localized artifacts~\cite{laa_net}. One solution to regularize training is data augmentation. As such, we propose, in addition to SBV, a novel Cutout data augmentation driven by vulnerable patches, i.e., image patches that are prone to blending artifacts~\cite{nguyen2024fakeformer}. We posit that by masking the most vulnerable regions, overfitting risks will be reduced, as the model will be pushed to learn from other areas. This masking strategy has already been explored in other computer vision fields such as Classification~\cite{normal_cutout, colorful_cutout}, Object Detection~\cite{cutout_objdet} demonstrating great potential.  

Specifically, similar to~\cite{fxray},  we create a set of blending boundaries $\mathbf B$ using a randomly generated blending mask $\mathbf M$ as follows,
\begin{equation}
      \mathbf B = (\mathbf 1 - \mathbf M) * \mathbf M * 4, \mathbf B \in \mathbb R^{T \times H \times W} \text{,}
\label{eq:blending_boundary} 
\end{equation} 

\noindent with $*$ being the element-wise multiplication and $\mathbf 1$ an all-one matrix. Inspired by~\cite{nguyen2024fakeformer}, vulnerability values are then quantified at the patch level in a non-overlapping manner by applying a MaxPooling function as follows,
\begin{equation}
    \bar{\mathbf B} = \text{MaxPooling}(\mathbf B), \hspace{2mm} \bar{\mathbf B} \in \mathbb R^{T \times \sqrt{N} \times \sqrt{N}} \text{,}
    \label{eq:vul_patch}
\end{equation}
\noindent where $N$ indicates the number of patches. 

After that, we define a threshold $\tau_{cutout}$ that is randomly selected from the range $(0.5, 1.0]$. We use the latter to define the set of patches to be masked $\mathcal P = \left\{ (l,m) \mid \bar{\mathbf B}_{l,m}(t_0) > \tau_{cutout} \right\}$ within the first frame. The set $\mathcal P$ is then used to mask out patches at those locations not only in the first frame but also in the entire video to enforce temporal consistency that is crucial for generating high-quality pseudo-fakes. After masking those patches over time, we finally obtain the masked blending boundary denoted as  $\tilde{\mathbf B}$. This results in masking the most vulnerable regions, i.e., the regions that are the most likely to include blending artifacts.  Figure~\ref{fig:overview_fw}-III shows some examples of the proposed cutout augmentation. 



\subsection{FakeSTormer}
\label{sec:fakestormer}


Our multi-task framework, called FakeSTormer, is inspired by~\cite{laa_net, nguyen2024fakeformer}, where auxiliary branches are designed to push the feature extractor to focus on vulnerabilities. As discussed earlier, the vulnerability is defined in~\cite{laa_net, nguyen2024fakeformer} as the pixels/patches that are the most likely to be impacted by blending artifacts. This strategy is therefore claimed to allow the detection of subtle artifacts that are generic across different types of manipulations. While such a vulnerability-driven approach has shown very promising results~\cite{laa_net, nguyen2024fakeformer}, it does not take into account the temporal nature of videos. Therefore, in addition to spatial vulnerabilities, we argue that there is a need to model temporal vulnerabilities, which we define as significant temporal changes in the blending boundary. 
Specifically, we introduce two additional branches, namely a temporal head $h$ and a spatial one $g$.  The branch $h$ predicts the derivatives of the blending boundary over time which can reflect high changes, typically characterizing temporal artifacts. Moreover,  we suggest the use of a spatial branch $g$ which enables predicting soft labels representing the forgery intensity encoded in each frame, computed from vulnerability information. 
The proposed framework relies on the TimeSformer backbone~\cite{timesformer}, which we revisit for better modeling spatial and temporal information.

Herein, we first describe the proposed revisited TimeSformer-based feature extractor $\Phi$ in Section~\ref{subsec:timesformer}. We then detail the two additional temporal and spatial heads in Section~\ref{subsec:temporal_modeling} and Section~\ref{subsec:spatial_sensibility}, respectively. Finally, we give the overall training details in Section~\ref{subsec:training_objective}.

\subsubsection{Backbone: Revisited TimeSformer}
\label{subsec:timesformer}
We choose TimeSformer~\cite{timesformer} as our feature extractor given its ability to effectively capture separate long-range temporal information and spatial features. In TimeSformer, a video input ${\mathbf X} \in \mathbb R^{C \times T \times H \times W}$ results in an embedding matrix input $\mathbf Z^0 \in \mathbb R^{T \times N \times D}$. A global class token $\mathbf z_{cls}$ attends all patches and is then used for classification.  This mechanism implicitly captures mixed spatio-temporal features, which might lead to overfitting one type of artifact. We revisit it slightly in order to decouple the spatial and temporal information by considering two sorts of additional tokens (one spatial and one temporal).

For that purpose, we attach in each dimension of $\mathbf Z^0$, a spatial token $\mathbf z_s^0 \in \mathbb R^D$ and a temporal token $\mathbf z_t^0 \in \mathbb R^D$, respectively. These tokens will independently interact only with patch embeddings belonging to their dimension axis by leveraging the decomposed SA~\cite{timesformer}. This mechanism not only facilitates the disentanglement learning process of spatio-temporal features but is also beneficial to optimize the computational complexity of $\mathcal O(T^2 + N^2)$ as compared to $\mathcal O(T^2 \cdot N^2)$ in vanilla SA. 
Those tokens will be then fed into $L$ ($L=12$ as default) transformer encoder blocks, as described in Figure~\ref{fig:overview_fw}-I. Formally, the feature extraction process can be summarized as follows,
\begin{equation}
    [\mathbf Z^L, \mathbf z^L_s, \mathbf z^L_t ] = \Phi( \mathbf X),
\end{equation}
\noindent where $\mathbf Z^L$ is the final patch embedding matrix, $\mathbf z^L_s$ the resulting set of spatial tokens, and $\mathbf z^L_t$ the resulting set of temporal tokens that will be respectively sent to the temporal head $h$, the spatial head $g$, and the classification head $f$. More details about the implementation of the proposed revisited TimeSformer are given in supplementary materials.

\subsubsection{Temporal Head $h$}\label{subsec:temporal_modeling}
\noindent\textbf{Ground Truths.} Our temporal head $h$ aims to model fine-grained temporal vulnerabilities in deepfake videos through a regression task.
First, to generate ground truth data for the branch $h$, we hypothesize that temporal high-changes in the blending boundary can reflect the presence of temporal artifacts (see Figure~\ref{fig:overview_fw}-IV). To achieve this, we compute $\mathbf D$ based on $\tilde{\mathbf B}$ such that:
\begin{align}
    \mathbf D &= \frac{\partial \tilde{\mathbf B}}{\partial t}, \hspace{2mm} \mathbf D \in \mathbb R^{ T\times \sqrt{N} \times \sqrt{N}}\label{equa:derivative} \text{.}
\end{align}
More details regarding the derivative calculation are provided in supplementary materials.
To stabilize training, $\mathbf D$ is standardized resulting in $ \hat{\mathbf D}  \in \mathbb R^{T \times \sqrt{N} \times \sqrt{N}}$. Experiments with different normalization strategies are reported in supplementary materials.

\vspace{1mm}
\noindent\textbf{Architecture Design.} In order to construct the regression head for predicting $ \hat{\mathbf D}$, we take the patch embedding  matrix $\mathbf Z^L$ as input and process them to produce 3D features as follows,
\begin{align}
    \mathbf F = \text{Reshape}(\mathbf Z^L), \mathbf F \in \mathbb R^{D \times T \times \sqrt{N} \times \sqrt{N}} \text{.}
\end{align}
To estimate temporal derivatives, we employ two 3D convolution blocks (3DCnvB) with $3$-dimensional temporal kernels and $1$-dimensional spatial kernels~\cite{ftcn} as follows,
\begin{equation}
    \tilde{\mathbf D} = h(\mathbf F)=\text{3DCnvB}_{3 \times 1 \times 1}(\text{3DCnvB}_{3 \times 1 \times 1}(\mathbf F)) \text{,}
\end{equation}
\noindent where $\tilde{\mathbf D} \in \mathbb R^{T \times \sqrt{N} \times \sqrt{N}}$. Each convolution block comprises a 3D convolution layer, followed by a BatchNorm and a GELU layer.

\vspace{1mm}
\noindent\textbf{Objective Function.} 
For training the temporal branch, we optimize the following Mean Squared Error (MSE) loss,
\begin{equation}
    \mathcal L_h = \frac{1}{T \times N} \|\hat{\mathbf D} - \tilde{\mathbf D} \|_2^2,
\end{equation}
with $ \|. \|_2$ referring to the $L_2$ norm.

\subsubsection{Spatial Head $g$}\label{subsec:spatial_sensibility}
\noindent\textbf{Ground Truths.} 
To avoid overfitting one type of artifact, we enforce the model to explicitly predict soft labels representing the intensity level of spatial artifacts for each video frame.  Note that several works~\cite{dinov1, deit, colla_learning, mixup, label_smoothing} have leveraged soft labels for training regularization.
Given a pseudo-fake video $\mathbf X=(\mathbf X(t))_{t \in [[1, T]]}$ formed by $T$ frames and $\tilde {\mathbf B}=(\tilde{\mathbf B}(t))_{t \in [[1, T]]}$ its associated cutout blending boundary, the ground truth for these soft labels is generated for each frame $t$ as follows,
\begin{equation}
    p(t) = \max_{l,m \in [[1,\sqrt{N}]]} (\tilde{\mathbf B}(t)) \text{,}
\end{equation}
resulting in the ground truth for training the spatial branch denoted as $\mathbf p = (p(t))_{t \in [[1,T]]} $. We note that $\mathbf p= \mathbf 1^T$ if cutout is not applied and $\mathbf p = \mathbf 0^T$ for a real video.

\vspace{1mm}
\noindent\textbf{Architecture Design.}
To predict the proposed soft labels, a Multi-Layer Perceptron (MLP) is applied to the set of spatial tokens $\mathbf z^L_s$, as follows,
\begin{equation}
    \tilde{\mathbf p} = g(\mathbf z^L_s)= \text{MLP}(\mathbf z^L_s), \hspace{2mm} \tilde{\mathbf p} \in \mathbb R^T.
\end{equation}

\vspace{1mm}
\noindent\textbf{Objective Function.}
To train the spatial branch, we optimize the following Binary Cross Entropy (BCE) loss similar to~\cite{mixup, deit},
\begin{equation}
    \mathcal L_g = \text{BCE}(\tilde{\mathbf p}, \mathbf p) \text{.}
\end{equation}





\subsubsection{Overall Training Objective}
\label{subsec:training_objective}

Finally, for the standard classification head $f$, we use the set of temporal tokens $\mathbf z^L_t$  such that the predicted label $\tilde y$ is given by,
\begin{equation}
    \tilde y=f(\mathbf z^L_t)= \text{MLP}(\mathbf z^L_t) \text{.}
\end{equation}
The classification loss $\mathcal L_c$ is then given by applying a BCE between the ground-truth label $y$ and the predicted label $\tilde y$.

Overall, the network is trained by optimizing the following loss:
\begin{equation}
    \mathcal L = \lambda_c\mathcal L_{c} + \lambda_h\mathcal L_h + \lambda_g\mathcal L_g \text{,}
\label{eq:total_loss}
\end{equation}
where $\lambda_c, \lambda_h, \lambda_g$ are hyper-parameters to balance the training of the three branches. 

\begin{table}[h]
\centering
\resizebox{\linewidth}{!}{
\begin{tabular}{c H cc H cccHHccc}
\hline
\multirow{2}{*}{Method} & \multirow{2}{*}{Inter.} & \multicolumn{2}{c}{Training} & \multirow{2}{*}{Pre.} & \multicolumn{8}{c}{Test set AUC (\%)} \\
\cline{3-4}
\cline{6-13}
& & Real & Fake & & CDF & DFD & DFDCP & DFo & FSh & DFDC & DFW & DiffSwap \\
\hline
\hline
Xception~\cite{ff++} & $\times$ & \checkmark & \checkmark & & 73.7 & - & - & 84.5 & 72.0 & 70.9 & - & - \\
MATT~\cite{multi-attentional} & $\times$ & \checkmark & \checkmark & & 68.3 & 92.9 & 63.0 & - & - & - & 65.7 & - \\
RECCE~\cite{ete_recons} & $\times$ & \checkmark & \checkmark & & 70.9 & \underline{98.2} & - & - & - & - & 68.2 & - \\
SBI~\cite{sbi} & $\times$ & \checkmark & $\times$ & & 90.6 & - & - & - & - & 72.4 & - & - \\
SFDG~\cite{sfdg} & $\times$ & \checkmark & \checkmark & & 75.8 & 88.0 & 73.6 & - & - & - & \underline{69.3} & - \\
LSDA~\cite{LSDA} & $\times$ & \checkmark & \checkmark & & \underline{91.1} & - & 77.0 & - & - & - & - & - \\

\hline
STIL~\cite{stil} & $\times$ & \checkmark & \checkmark & \checkmark & 75.6 & - & - & - & - & - & - & - \\
LipForensics~\cite{lipforensics} & $\times$ & \checkmark & \checkmark & \checkmark & 82.4 & - & - & 97.6 & 97.1 & \underline{73.5} & - & - \\
RealForensics~\cite{realforensics} & $\times$ & \checkmark & \checkmark & \checkmark & 86.9 & 82.2 & 75.9 & 99.3 & 99.7 & - & - & - \\
FTCN~\cite{ftcn} & $\times$ & \checkmark & \checkmark & \checkmark & 86.9 & 94.4 & 74.0 & 98.8 & 98.8 & 71.0 & - & - \\
ISTVT~\cite{istvt} & $\times$ & \checkmark & \checkmark & \checkmark & 84.1 & - & 74.2 & - & - & - & - & - \\
AltFreezing~\cite{altfreezing} & $\times$ & \checkmark & \checkmark & \checkmark & 89.5 & \textbf{98.5} & - & 99.3 & 99.4 & - & - & - \\
Swin+TALL~\cite{tall_swin} & $\times$ & \checkmark & \checkmark & \checkmark & 90.8 & - & 76.8 & 99.2 & 99.7 & - & - & - \\
StyleLatentFlows~\cite{stylelatent} & $\times$ & \checkmark & \checkmark & \checkmark & 89.0 & 96.1 & - & 99.0 & 99.2 & - & - & - \\
LFGDIN~\cite{LFGDIN} & $\times$ & \checkmark & \checkmark & \checkmark & 90.4 & - & \underline{80.8} & - & - & - & - & \underline{85.7} \\



\hline
\hline
FakeSTormer ($T=4$) & \checkmark & \checkmark & $\times$ & \checkmark & \textbf{92.4} & \textbf{98.5} & \textbf{90.0} & - & 84.1 & \textbf{74.6} & \textbf{74.2} & \textbf{96.9} \\

FakeSTormer ($T=8$) & \checkmark & \checkmark & $\times$ & \checkmark & 92.4 & 98.2 & 90.0 & - & - & 74.9 & 75.9 & 97.1 \\

FakeSTormer ($T=16$) & \checkmark & \checkmark & $\times$ & \checkmark & 92.8 & 98.6 & 90.2 & - & - & 75.1 & 75.3 & 97.2 \\
\hline
\end{tabular}%
}
\vspace{-3mm}
\caption{\textbf{Generalization to unseen datasets}. AUC (\%) comparisons at \textit{video-level} on multiple unseen datasets~\cite{celeb_df, dfd, dfdcp, dfdc, wdf, DiffSwap}. All detectors are trained on FF++(c23). Results are directly extracted from the original papers and from~\cite{laa_net, lipforensics}. \textbf{Bold} and \underline{Underlined} text, respectively highlight the best and the second best performance, excluding the variants of our framework with $T=8$ and $T=16$.}
\vspace{-4mm}
\label{tabl:cross_dataset_eval}
\end{table}






\section{Experiment}
\label{sec:exp}

\begin{table}
\centering
\resizebox{\linewidth}{!}{
\begin{tabular}{c H cc H ccc H ccc}
\hline
\multirow{2}{*}{Method} & \multirow{2}{*}{Inter.} & \multicolumn{2}{c}{Training set} & & \multicolumn{3}{c}{Cross-dataset} & & \multicolumn{3}{c}{DF40 subset} \\
\cline{3-4}
\cline{6-8}
\cline{10-12}
& & Real & Fake & & CDF & DFDCP & DiffSwap & & BlendFace & FSGAN & MobileSwap \\
\hline
\hline
Face X-ray~\cite{fxray} & \checkmark & \checkmark & \checkmark & & 79.5 & - & - & - & - & - & - \\
PCL+I2G~\cite{cstency_learning} & $\times$ & \checkmark & \checkmark & & 90.0 & 74.3 & - & - & - & - & - \\
SLADD~\cite{sladd} & \checkmark & \checkmark & \checkmark & & 79.7 & - & - & - & - & - & - \\
SBI~\cite{sbi} & $\times$ & \checkmark & $\times$ & & 93.2 & 86.2 & 90.6 & - & 86.5 & 85.4 & 86.6 \\
LAA-Net~\cite{laa_net} & \checkmark & \checkmark & $\times$ & & 95.4 & 86.9 & 92.1 & - & \textbf{91.2} & 94.2 & 93.9 \\

\hline
STC-Scratch~\cite{STC} & $\times$ & \checkmark & $\times$ & & 83.4 & 86.8 & - & - & - & - & - \\
STC-Pretrain~\cite{STC} & $\times$ & \checkmark & $\times$ & & \underline{95.8} & 89.4 & - & - & - & - & - \\
ST-SBV~\cite{stsbv} & $\times$ & \checkmark & $\times$ & & 90.3 & 91.2 & - & - & - & - & - \\
StA+VB~\cite{plug_play} & $\times$ & \checkmark & $\times$ & & 94.7 & 90.9 & - & - & 90.6 & \textbf{96.4} & \underline{94.6} \\

\hline
\hline
TimeSformer~\cite{timesformer} + SBV & $\times$ & \checkmark & $\times$ & & 94.9 & \underline{93.0} & \underline{93.3} & - & 89.7 & \underline{94.6} & \underline{94.6} \\
\hline
FakeSTormer & \checkmark & \checkmark & $\times$ & & \textbf{96.5} & \textbf{94.1} & \textbf{97.7} & - & \underline{91.1} & \textbf{96.4} & \textbf{95.0} \\
\hline
\end{tabular}%
}
\vspace{-3mm}
\caption{\textbf{AUC(\%) comparison at video-level with other data synthesis methods.} For fair comparison, we train our FakeSTormer on raw data of FF++(c0), and test \textit{cross-dataset} on~\cite{celeb_df, dfdcp, DiffSwap} and \textit{cross-manipulation} on three subsets of~\cite{DF40}.}
\vspace{-4mm}
\label{tabl:data_synthesis_comp}
\end{table}

\subsection{Settings}
\label{sec:setup}

\noindent\textbf{Datasets.} 
We set up our datasets following several works~\cite{altfreezing, ftcn, tall_swin, istvt, stylelatent, ucf, sfdg, ete_recons}. For both training and validation, we employ \textbf{FaceForensics++} (FF++)~\cite{ff++}, which consists of four manipulation methods for the fake data (Deepfakes (DF)~\cite{deepfake}, FaceSwap (FS)~\cite{faceswap}, Face2Face (F2F)~\cite{face2face}, and NeuralTextures (NT)~\cite{neutex}). It can be noted that, for training, we use only the real videos and generate pseudo-fake data using our synthesized method, SBV. \textit{By default, the c23 version of FF++ is adopted}, following the recent literature~\cite{ftcn, altfreezing, istvt, tall_swin, stylelatent}. For further validation, we also evaluate on the following datasets: \textbf{Celeb-DFv2} (CDF)~\cite{celeb_df}, \textbf{DeepfakeDetection} (DFD)~\cite{dfd}, \textbf{Deepfake Detection Challenge Preview} (DFDCP)~\cite{dfdcp}, \textbf{Deepfake Detection Challenge} (DFDC)~\cite{dfdc}, \textbf{WildDeepfake} (DFW)~\cite{wdf}, \textbf{DF40}~\cite{DF40}, and \textbf{DiffSwap}~\cite{LFGDIN, DiffSwap} generated using a recent diffusion-based approach~\cite{DiffSwap}. Further details on these datasets are provided in supplementary materials.

\begin{table}
\centering
\scalebox{0.7}{
\begin{tabular}{c cc H cc}
\hline
\multirow{2}{*}{Method} & \multicolumn{2}{c}{Training set} & & \multicolumn{2}{c}{FF++ LQ (\%)} \\
\cline{2-3}
\cline{5-6}
& Real & NT & 
& DF & FS \\
\hline
\hline
Xception~\cite{ff++} & \checkmark & \checkmark & 
& 58.7 & 51.7 \\
Face X-ray~\cite{fxray} & \checkmark & \checkmark & 
& 57.1 & 51.0 \\
F3Net~\cite{f3net} & \checkmark & \checkmark & 
& 58.3 & 51.9 \\
RFM~\cite{rfm} & \checkmark & \checkmark & 
& 55.8 & 51.6 \\
SRM~\cite{srm} & \checkmark & \checkmark & 
& 55.5 & 52.9 \\
SLADD~\cite{sladd} & \checkmark & \checkmark & 
& 62.8 & 56.8 \\
\hline
TALL-Swin~\cite{tall_swin} & \checkmark & \checkmark & 
& 63.2 & 51.4 \\
ResNet3D$^\ast$~\cite{Resnet3D} & \checkmark & \checkmark & 
& 66.8 & \underline{60.6} \\
TimeSformer$^\ast$~\cite{timesformer} & \checkmark & \checkmark & 
& \underline{73.3} & 54.4 \\
\hline
\hline
Ours & \checkmark & $\times$ & 
& \textbf{85.3} & \textbf{62.1} \\
\hline
\end{tabular}%
}
\vspace{-3mm}
\caption{\textbf{Generalization on heavily compressed data (LQ).} AUC (\%) comparisons on FF++ (LQ)~\cite{ff++} with a high compression level (c40). The results for comparison are directly extracted from~\cite{sladd, domainforensic}. The symbol $\ast$ denotes our implementation.}
\vspace{-4mm}
\label{tabl:cross_mani_eval}
\end{table}

\noindent\textbf{Data Pre-processing.} Following the splitting convention~\cite{ff++}, we extract $256$, $32$, and $32$ consecutive frames for training, validation, and testing, respectively. Facial regions are cropped using Face-RetinaNet~\cite{retina_face} and resized to a fixed resolution of $224 \times 224$. Additionally, we store $81$ facial landmarks for each frame, extracted using Dlib~\cite{dlib}. Further details are provided in the supplementary materials.

\noindent\textbf{Evaluation Metrics.} For fair comparisons with SOTA methods, we use the widely adopted Area Under the Curve (AUC) metric at the video level~\cite{altfreezing, stylelatent, ftcn, lipforensics, realforensics, tall_swin, istvt}.

\begin{figure*}
    \centering
    \includegraphics[width=0.85\linewidth]{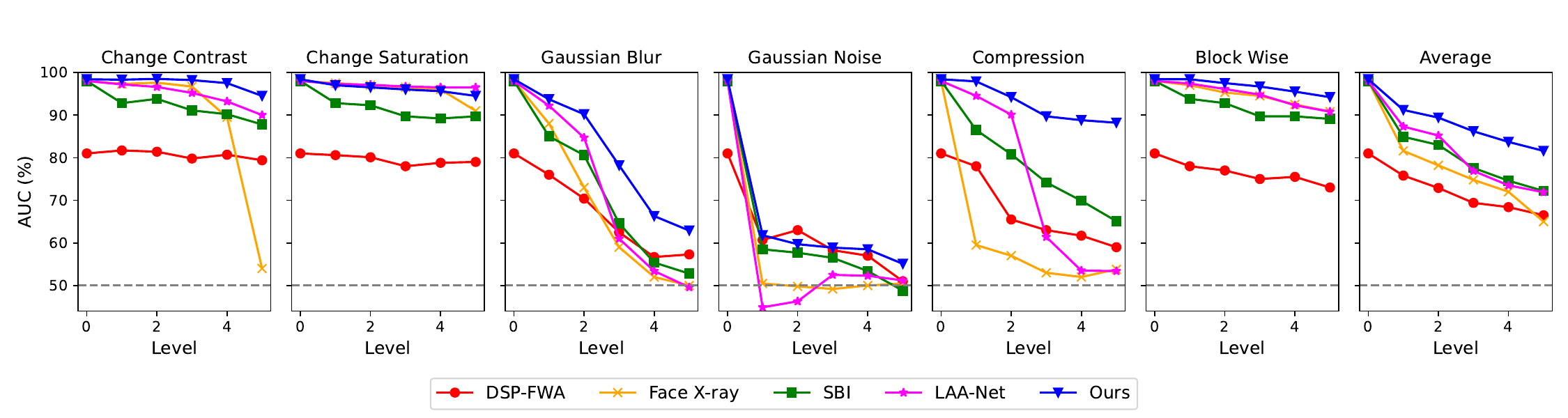}
    \vspace{-3mm}
    \caption{\textbf{Robustness to unseen perturbations.} AUC (\%) under five different degradation levels for various types of perturbations~\cite{DFo} on FF++~\cite{ff++}.  ``Average'' denotes the mean across all corruptions at each level. Best viewed in color.}
    \vspace{-4mm}
    \label{fig:robustness}
\end{figure*}

\noindent\textbf{Implementation Details.} 
Our framework is initialized with pretrained MAE weights~\cite{mae} and trained for $100$ epochs using the SAM optimizer~\cite{sam} with a weight decay of $10^{-4}$ and a batch size of $32$. The learning rate starts at $5 \times 10^{-4}$ for the first quarter of training and decays to $0$ thereafter. The backbone is frozen for the first $5$ epochs for warm-up, then all layers are unfrozen. Data augmentation includes ColorJittering at the video level and our proposed Cutout. Experiments are conducted on four NVIDIA A100 GPUs, with $\tau = 0.35$ and $\bar{d} = 0.2$ (Eq.~\eqref{eq:lmk_interpolation}), and $T = 4$ frames in most experiments.

\begin{table}
\centering
\scalebox{0.7}{
\begin{tabular}{c cc H ccccc}
\hline
\multirow{2}{*}{Method} & \multicolumn{2}{c}{Training set} & & \multicolumn{5}{c}{FF++ (\%)} \\
\cline{2-3}
\cline{5-8}
& Real & Fake & & DF & FS & F2F & NT & Avg. \\
\hline
\hline
Xception~\cite{ff++} & \checkmark & \checkmark & & 93.9 & 51.2 & 86.8 & 79.7 & 77.9 \\
Face X-ray~\cite{fxray} & \checkmark & \checkmark & & 99.5 & 93.2 & 94.5 & 92.5 & 94.9 \\
SBI~\cite{sbi} & \checkmark & $\times$ & & 98.6 & 95.4 & 92.6 & 82.3 & 92.2 \\
LSDA~\cite{LSDA} & \checkmark & \checkmark & & 96.9 & 95.1 & 96.4 & 94.9 & 95.8 \\

\hline
LipForensics~\cite{lipforensics} & \checkmark & \checkmark & & 99.7 & 90.1 & 99.7 & 99.1 & 97.1 \\
FTCN~\cite{ftcn} & \checkmark & \checkmark & & 99.8 & 99.6 & 98.2 & 95.6 & 98.3 \\
RealForensics~\cite{realforensics} & \checkmark & \checkmark & & \textbf{100} & 97.1 & 99.7 & 99.2 & 99.0 \\
AltFreezing~\cite{altfreezing} & \checkmark & \checkmark & & 99.8 & \underline{99.7} & 98.6 & 96.2 & 98.6 \\
StyleLatentFlows~\cite{stylelatent} & \checkmark & \checkmark & & 99.7 & 98.8 & 98.6 & 96.4 & 98.4 \\
NACO~\cite{NACO} & \checkmark & \checkmark & & \underline{99.9} & \underline{99.7} & \underline{99.8} & \underline{99.4} & \underline{99.7} \\
LFGDIN~\cite{LFGDIN} & \checkmark & \checkmark & & 96.2 & 80.5 & 90.5 & 81.7 & 87.2 \\

\hline
\hline
Ours (c23) & \checkmark & $\times$ & & \underline{99.9} & 97.8 & 98.5 & 97.2 & 98.4 \\
Ours (c0) & \checkmark & $\times$ & & \textbf{100} & \textbf{99.8} & \textbf{99.9} & \textbf{99.7} & \textbf{99.9} \\
\hline
\end{tabular}%
}
\vspace{-3mm}
\caption{\textbf{Generalization to unseen manipulations}. AUC (\%) comparisons on FF++~\cite{ff++}, which consists of four manipulation methods (DF, FS, F2F, NT). 
}
\vspace{-4mm}
\label{tabl:indataset_eval}
\end{table}

\subsection{Comparison with State-of-the-art Methods}
\label{sec:compare_sota}

\noindent\textbf{Generalization to Unseen Datasets.} 
To assess the generalization capabilities of our method, we conduct evaluations using the challenging \textit{cross-dataset} setup~\cite{altfreezing, ftcn, laa_net, sfdg, ete_recons}, validating on unseen datasets (i.e., datasets other than FF++). The results are detailed in Table~\ref{tabl:cross_dataset_eval} and Table~\ref{tabl:data_synthesis_comp}.


As shown, our method achieves comparable results on DFD while surpassing SOTA methods on other datasets. Specifically, it significantly outperforms prior video deepfake detection techniques, including spatio-temporal learning-based methods like AltFreezing~\cite{altfreezing} and ISTVT~\cite{istvt}, as well as various data synthesis approaches. Moreover, our method exhibits superior performance on the large-scale DFDC dataset and the challenging in-the-wild DFW dataset. These results further confirm the enhanced generalization ability of FakeSTormer compared to recent methods.

\begin{table}
\centering
\resizebox{\linewidth}{!}{
\begin{tabular}{cccc ccccccc}
\hline
\multirow{2}{*}{SBV} & \multirow{2}{*}{V-CutOut} & \multirow{2}{*}{$g$}& \multirow{2}{*}{$h$} & \multicolumn{7}{c}{Test set AUC (\%)} \\
\cline{5-10}
& & & & CDF & DFD & DFDCP & DFDC & DFW & DiffSwap & Avg. \\
\hline
\hline
$\times$ & $\times$ & $\times$ & $\times$ & \textcolor{gray}{61.5} & \textcolor{gray}{62.8} & \textcolor{gray}{59.4} & \textcolor{gray}{58.5} & \textcolor{gray}{65.2} & \textcolor{gray}{71.6} & \textcolor{gray}{63.2} \\

\checkmark & $\times$ & $\times$ & $\times$ & 90.7 & 95.7 & 87.9 & 72.2 & 70.9 & 92.9 & 85.1(\textcolor{ForestGreen}{$\uparrow$21.9}) \\

\checkmark & \checkmark & $\times$ & $\times$ & 91.1 & \underline{96.0} & 87.6 & 72.6 & 71.0 & 93.1 & 85.2(\textcolor{ForestGreen}{$\uparrow$22.0}) \\

\checkmark & \checkmark & \checkmark & $\times$ & 92.2 & 95.4 & \underline{88.5} & \underline{72.8} & \underline{71.3} & 93.8 & 85.7(\textcolor{ForestGreen}{$\uparrow$22.5}) \\

\checkmark & $\times$ & $\times$ & \checkmark & \textbf{93.4} & \textbf{98.5} & \underline{88.5} & \underline{72.8} & 69.6 & \textbf{97.3} & \underline{86.7}(\textcolor{ForestGreen}{$\uparrow$23.5}) \\

\checkmark & \checkmark & \checkmark & \checkmark & \underline{92.4} & \textbf{98.5} & \textbf{90.0} & \textbf{74.6} & \textbf{74.2} & \underline{96.9} & \textbf{87.8}(\textcolor{ForestGreen}{$\uparrow$24.6}) \\
\hline
\end{tabular}%
}
\vspace{-3mm}
\caption{\textbf{Ablation study of framework's components}. \textcolor{gray}{Gray} indicates the use of original fake data for training.}
\vspace{-4mm}
\label{tabl:comp_abl}
\end{table}

\noindent\textbf{Generalization on Heavily Compressed Data.} 
Following previous work~\cite{domainforensic, sladd}, we also evaluate FakeSTormer on heavily compressed FF++(c40) data. In addition to comparing with several SOTA methods, we train ResNet3D~\cite{Resnet3D}, commonly used in deepfake video detection~\cite{ftcn, altfreezing, stylelatent}, and TimeSformer~\cite{timesformer} on NT, then test on DF and FS. The comparison results are presented in Table~\ref{tabl:cross_mani_eval}. Our method achieves notably higher AUC scores than other methods across both testing subsets, highlighting its robust generalization capability under various data compression conditions.

\noindent\textbf{Generalization to Unseen Manipulations.} 
Table~\ref{tabl:indataset_eval} compares our framework with SOTA methods on FF++. Other methods~\cite{lipforensics, altfreezing, stylelatent, ftcn} use a cross-manipulation setup, training on three forgery types and evaluating on the remaining one. In contrast, our approach trains only on real videos, treating all manipulations as unseen. Despite this, our method shows competitive performance with the others, even without being trained on specific forgery types.

\noindent\textbf{Robustness to Unseen Perturbations.} 
Deepfakes are widely shared on social media, where various perturbations can affect their appearance. Following~\cite{DFo}, we evaluate FakeSTormer's robustness across six unseen degradation types at five levels, comparing it with other augmented-based methods~\cite{fwa, fxray, sbi, laa_net}. Figure~\ref{fig:robustness} shows AUC scores for each method on these perturbations, using models trained on FF++. Our results demonstrate that FakeSTormer outperforms prior methods on most distortions, with a slight drop compared to LAA-Net~\cite{laa_net} for Change Saturation. Nonetheless, FakeSTormer achieves higher performance on average, especially at higher severity levels, highlighting its superior generalization and robustness. Detailed scores are in supplementary materials.


\subsection{Additional Discussions}
\label{sec:abl}

\noindent\textbf{Ablation Study of the FakeSTormer's Components.}
We conduct ablation studies to assess the impact of each component in our framework, as shown in Table~\ref{tabl:comp_abl}. Using TimeSformer trained on FF++ as the baseline, we experiment with different combinations of components: Self-Blended Video (SBV), Vulnerability-driven CutOut (V-CutOut), the spatial branch ($g$), and the temporal branch ($h$). Each component improves performance, with SBV providing the most significant boost by generating high-quality pseudo-fake data that aids generalization to unseen datasets.


\noindent\textbf{Ablation Study of the SBV's Components.} 
SBV enhances SBI~\cite{sbi} with CSP and LI for robust pseudo-fake generation in video data. Table~\ref{tabl:sbv_abl} shows that without these components, simply stacking frame-wise SBIs fails to produce consistent temporal features, leading to overfitting on more obvious artifacts and poor generalization~\cite{altfreezing}. A qualitative comparison is provided in supplementary materials.

\begin{table}
\centering
\resizebox{\linewidth}{!}{
\begin{tabular}{c c c c cccccc}
\hline
\multirow{2}{*}{Method} & \multirow{2}{*}{Comp.} & \multirow{2}{*}{CSP} & \multirow{2}{*}{LI} & \multicolumn{6}{c}{Test set AUC (\%)} \\
\cline{5-10}
 & & & & CDF & DFD & DFDCP & DFDC & DFW & DiffSwap \\
\hline
\hline
Stacked SBIs & c23 & $\times$ & $\times$ & 48.4 & 49.1 & 48.9 & 51.7 & 52.8 & 55.5 \\
+ CSP & c23 & \checkmark & $\times$ & 84.1 & 89.2 & 86.1 & 69.5 & 65.5 & 85.7 \\
SBV & c23 & \checkmark & \checkmark & \underline{90.7} & \underline{95.7} & \underline{87.9} & \underline{72.2} & \underline{70.9} & \underline{92.9} \\
SBV & c0 & \checkmark & \checkmark & \textbf{94.9} & \textbf{97.6} & \textbf{93.0} & \textbf{76.4} & \textbf{75.3} & \textbf{93.3} \\
\hline
\end{tabular}%
}
\vspace{-3mm}
\caption{\textbf{Ablation study of SBV's components}. Performance analyses of different SBV's components using cross-evaluation on multiple datasets~\cite{celeb_df, dfd, dfdcp, dfdc, wdf, DiffSwap}.}
\vspace{-4mm}
\label{tabl:sbv_abl}
\end{table}

\noindent\textbf{Influence of Number of Frames.} 
Increasing the number of frames $T$ provides more fine-grained temporal information. In Table~\ref{tabl:cross_dataset_eval}, we vary $T$ values by fixing it to $4$, $8$, and $16$. Our results show a consistent performance improvement with more frames, confirming our hypothesis. However, increasing $T$ also incurs a higher computational cost.



\begin{table}
\centering
\resizebox{\linewidth}{!}{
\begin{tabular}{ccc ccccccc}
\hline
\multirow{2}{*}{$\lambda_c$} & \multirow{2}{*}{$\lambda_h$} & \multirow{2}{*}{$\lambda_g$} & \multicolumn{7}{c}{Test set AUC (\%)} \\
\cline{4-9}
& & & CDF & DFD & DFDCP & DFDC & DFW & DiffSwap & Avg. \\
\hline
\hline
0.9 & 1 & 0.1 & 89.5 & 91.0 & \textbf{93.2} & 71.5 & \textbf{74.7} & 90.6 & 85.1 \\
0.9 & 10 & 0.1 & \textbf{92.5} & 95.7 & 86.8 & 71.7 & 72.7 & 94.5 & 85.7 \\
0.9 & 100 & 0.1 & 91.6 & \underline{98.0} & 87.3 & 73.6 & 70.6 & 96.2 & 86.2 \\
0.8 & 100 & 0.2 & \underline{92.4} & \textbf{98.5} & \underline{90.0} & \textbf{74.6} & \underline{74.2} & \underline{96.9} & \textbf{87.8} \\
0.5 & 100 & 0.5 & \underline{92.4} & \underline{98.0} & 88.1 & \underline{74.5} & 72.1 & \textbf{97.1} & \underline{87.0} \\
\hline
\end{tabular}%
}
\vspace{-3mm}
\caption{\textbf{Impact of loss balancing factors}. AUC (\%) comparisons of FakeSTormer trained with different values of $\lambda_c$, $\lambda_h$, and $\lambda_g$ on cross-dataset setup, demonstrating robustness to varying hyperparameter settings.}
\vspace{-4mm}
\label{tabl:lambda_abl}
\end{table}

\noindent\textbf{Impact of Loss Balancing Factors.} We introduce three hyperparameters, $\lambda_c$, $\lambda_g$, and $\lambda_h$ in Eq~\eqref{eq:total_loss} to balance the training among the three branches of our framework. In Table~\ref{tabl:lambda_abl}, we analyze the impact of these hyperparameters using various values. Our results show that the method is robust to a range of hyperparameter values, with the best performance achieved when $\lambda_c$, $\lambda_g$, and $\lambda_h$ are set to $0.8$, $0.2$, and $100$, respectively.

\subsection{Visualization of Saliency Maps}
\begin{figure}
    \centering
    \includegraphics[width=0.95\linewidth]{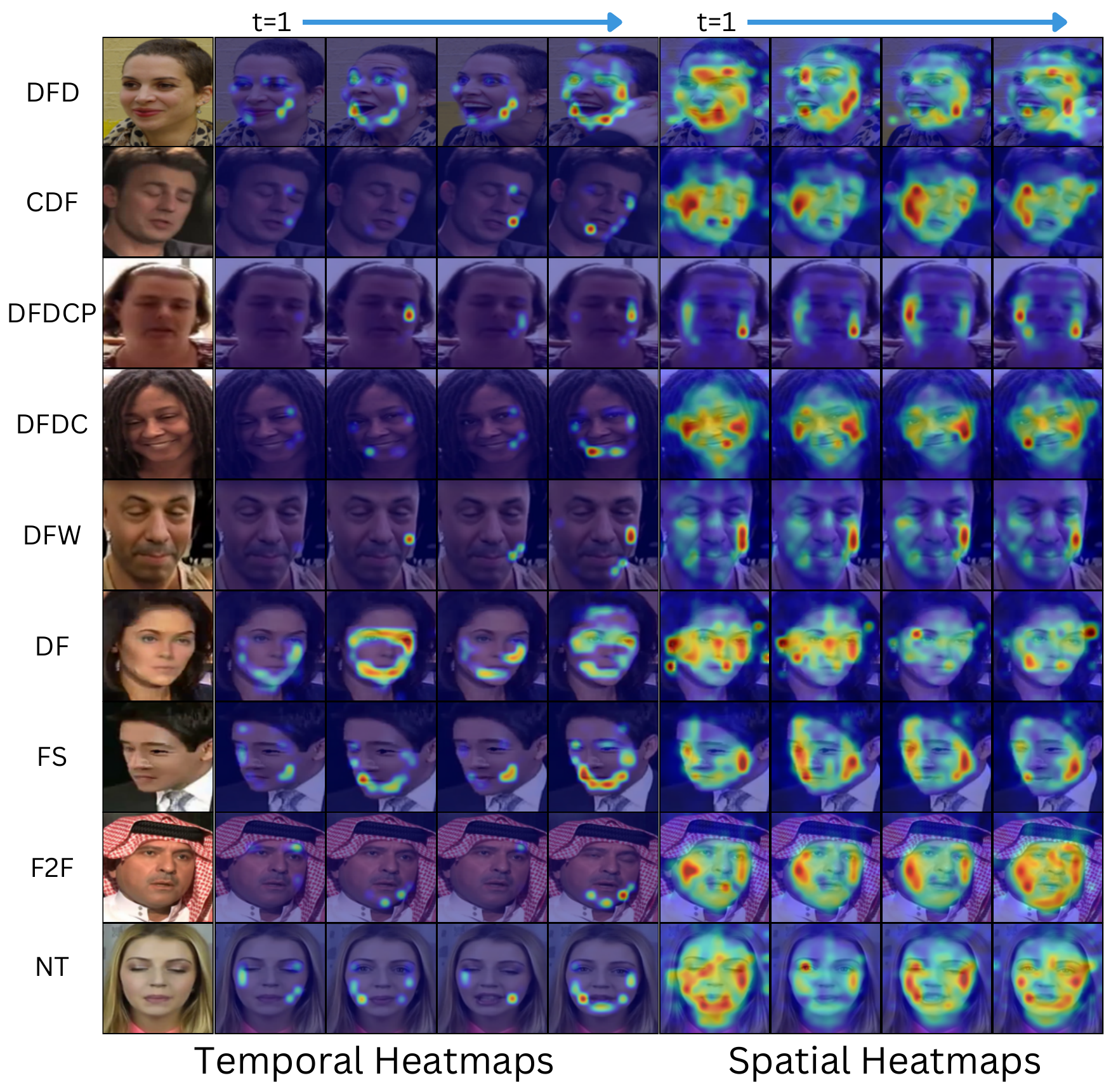}
    \vspace{-2mm}
    \caption{\textbf{Visualization of Saliency Maps}. The second-fifth and sixth-ninth columns represent temporal heatmaps and spatial heatmaps on different frames in the video, respectively. All datasets are unseen during validation.}
    \label{fig:qualitative_res}
    \vspace{-5mm}
\end{figure}

To analyze the contribution of the two proposed branches $h$ and $g$ in the detection performance of FakeSTormer, we visualize the input regions activated by those branches. For that purpose, we adopt Grad-CAM~\cite{gradCAM} for the temporal branch $h$ and utilize the final SA scores of spatial tokens for the spatial branch $g$. The visualization results from various datasets are presented in Figure~\ref{fig:qualitative_res}. It can be observed that FakeSTormer can discriminate between real and fake videos by focusing on very few, different local areas, even without having seen those types of forgeries during training.

\section{Conclusion}
\label{sec:conclu}

This paper introduces a fine-grained approach for generalizable deepfake video detection with two main contributions. First, we propose a multi-task learning framework that targets both subtle spatial and fine-grained temporal vulnerabilities in high-fidelity deepfake videos, incorporating a standard classification branch along with two new auxiliary branches (temporal and spatial). These proposed branches help the model focus on vulnerable regions and provide more valuable insights into how the network \textit{sees} the data while offering more robustness to high-quality deepfakes. This framework is further supported by the introduction of a high-quality pseudo-fake generation technique. Extensive experiments on several challenging benchmarks demonstrate that FakeSTormer achieves superior performance compared to SOTA methods.

\vspace{3mm}
\noindent\textbf{\large{Acknowledgment}}
\label{sec:acknowledge}
\vspace{1mm}

This work is supported by the Luxembourg National Research Fund, under the BRIDGES2021/IS/16353350/FaKeDeTeR, and by POST Luxembourg. Experiments were performed on the Luxembourg national supercomputer MeluXina. The authors gratefully acknowledge the LuxProvide teams for their expert support.

{
    \small
    \bibliographystyle{ieeenat_fullname}
    \bibliography{main}
}

\clearpage
\setcounter{page}{1}
\maketitlesupplementary

\section{Appendix}


\subsection{Details of the Ground-Truth Derivative}

\noindent\textbf{Calculation formula.}
To generate the ground truth data for the temporal branch $h$, we compute the derivative on $\Tilde{\mathbf B} = (\Tilde{\mathbf B}(t))_{t \in [[1, T]]}$ with respect to the temporal dimension. Specifically, we calculate the absolute value of the difference between two consecutive patch-level vulnerability values $\Tilde{\mathbf B}(t)$ and $\Tilde{\mathbf B}(t-1)$ with $t \geq 2$ such that,

\begin{equation}
    \mathbf D(t) = \left| \Tilde{\mathbf B}(t) - \Tilde{\mathbf B}(t-1) \right|.
\end{equation}

The process is iterated for every pair of consecutive frames of $(\Tilde{\mathbf B}(t))_{t \in [[1, T]]}$ to obtain a derivative matrix $\mathbf D \in \mathbb R^{T \times \sqrt{N} \times \sqrt{N}}$. 
For $t=1$, we insert a matrix $\mathbf 0$, i.e., $\mathbf D(1) = \mathbf 0$, indicating no temporal change at the first frame.

\noindent\textbf{Normalization functions.} Employing a normalization function is important for stabilizing the training of our temporal branch $h$. 
Therefore, we consider three different normalization functions including Standardization (MeanStd), MinMax, and Unnormalized 3D Gaussian.
Specifically, for the Standardization and MinMax, we respectively compute the std($\mathbf D$)-mean($\mathbf D$), and min($\mathbf D$)-max($\mathbf D$), while we follow the work of~\cite{nguyen2024fakeformer} to adapt an unnormalized Gaussian map from 2D to 3D for normalization.
We report in Table.~\ref{tabl:norm_func_abl} cross-evaluation results on six datasets~\cite{celeb_df, dfd, dfdcp, dfdc, wdf, DiffSwap} with the use of the three investigated functions, using a model trained on FF++~\cite{ff++}.
It can be noted that the model is robust to various types of normalization functions with the best performance recorded for the Standardization approach.

\begin{table}
\centering
\resizebox{\linewidth}{!}{
\begin{tabular}{c ccccccc}
\hline
\multirow{2}{*}{Function} & \multicolumn{7}{c}{Test set AUC (\%)} \\
\cline{2-7}
 & CDF & DFD & DFDCP & DFDC & DFW & DiffSwap & Avg. \\
\hline
\hline
MinMax & 91.6 & \underline{95.9} & \underline{88.1} & 71.1 & 70.1 & \underline{92.1} & 84.8 \\
Unnormalized 3D Gaussian & \textbf{92.5} & 91.8 & 86.4 & \underline{72.4} & \textbf{76.3} & 91.7 & \underline{85.2} \\
MeanStd & \underline{92.4} & \textbf{98.5} & \textbf{90.0} & \textbf{74.6} & \underline{74.2} & \textbf{96.9} & \textbf{87.8} \\
\hline
\end{tabular}%
}
\caption{\textbf{Comparison of different normalization functions}. We consider three normalization functions, i.e., Standardization (MeanStd), MinMax, and Unnormalized 3D Gaussian. Among these, Standardization gives the best overall performance.}
\label{tabl:norm_func_abl}
\end{table}

\subsection{SBV: Pseudo-code and Visual Samples}
\label{sec:sbv_samples}
\noindent\textbf{Algorithm.} To enhance the clarity and reproducibility of the SBV generation process, we provide the overall algorithm in the form of pseudo-code, as detailed in Algorithm~\ref{alg:sbvs}.

\begin{figure*}
    \centering
    \includegraphics[width=\linewidth]{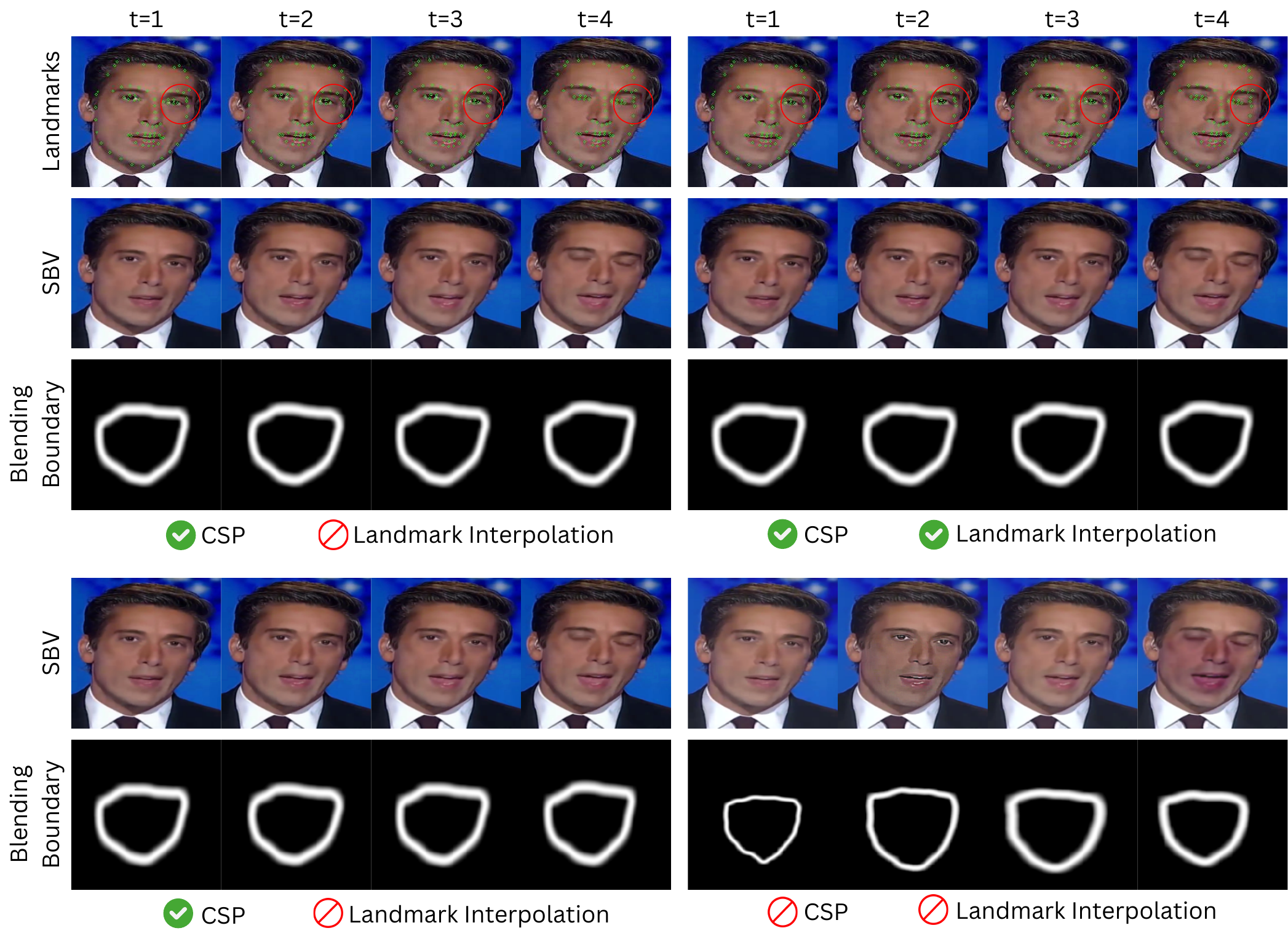}
    \caption{\textbf{Illustration of the facial landmarks, the generated SBV, and the blending boundaries} with and without applying the Consistent Synthesized Parameters (CSP) module and the Landmark Interpolation (LI) module. The lack of applied CSP and LI indicates simply stacked SBIs (BottomRight).}
    \label{fig:lmk_interpolation}
\end{figure*}

\begin{algorithm}[]
   \SetAlgoLined
   \KwIn{Real video $\mathbf X^r_i \in \mathcal V^r$ of size $(C,T,H,W)$, facial landmarks $\mathbf L_i \triangleq \cup_{t=1}^T \{\mathbf l_{ij} (t)\}_{1 \leq j \leq n}$ of size $(T,n,2)$, a distance $\bar d$, a threshold $\tau$}
   \KwOut{Self-blended video $\mathbf X^{\Tilde r}_i \in \mathcal V^{\Tilde r}$ of size $(C,T,H,W)$, blending mask $\mathbf M_i$ of size $(T,H,W)$}
    Initialize $\theta^{(sbi)}$ as an empty dictionary \\
    Initialize $\mathbf X_i^{\Tilde{r}}$ as an empty array \\
    Initialize $\mathbf M_i$ as an empty array \\
    \For{$j = 1$ \KwTo $T$}{
       \If{$j = 1$}{
           $\mathbf X^{\Tilde r}_i (t_0)$, $\mathbf M_i (t_0)$, \{$\theta^{(c)}$, $\theta^{(m)}$, $\theta^{(b)}$, $\dots$\} $\leftarrow$ SBI($\mathbf X_i^r (t_0)$, $\mathbf l_i (t_0)$) \\ 
           $\mathbf X_i^{\Tilde{r}} \gets \mathbf X_i^{\Tilde{r}} \cup \{\mathbf X^{\Tilde r}_i (t_0)\}$ \\
           $\mathbf M_i \gets \mathbf M_i \cup \{\mathbf M_i (t_0)\}$ \\
           $\theta^{(sbi)}$ $\leftarrow$ \{$\theta^{(c)}$, $\theta^{(m)}$, $\theta^{(b)}$, \dots\} \\
        }
        \Else{
           $\mathbf l_i (t)$ $\leftarrow$ LandmarkInterpolation($\mathbf l_i (t)$, $\mathbf l_i (t-1)$, $\bar d$, $\tau$) \\
           $\mathbf X^{\Tilde r}_i (t)$, $\mathbf M_i (t)$, $\_$ $\leftarrow$ SBI($\mathbf X_i^r (t)$, $\mathbf l_i (t)$, $\theta^{(sbi)}$) \\
           $\mathbf X_i^{\Tilde{r}} \gets \mathbf X_i^{\Tilde{r}} \cup \{\mathbf X^{\Tilde r}_i (t)\}$ \\
            $\mathbf M_i \gets \mathbf M_i \cup \{\mathbf M_i (t)\}$
        }
    }
    \Return{$\mathbf X_i^{\Tilde r}$, $\mathbf M_i$}
    \caption{Pseudo-code for SBV Generation}
    \label{alg:sbvs}
\end{algorithm}

\noindent\textbf{Visual Samples.}
To visually demonstrate the benefits of the Consistent Synthesized Parameters (CSP) and the Landmark Interpolation (LI) module (Section~\textcolor{red}{3.1} in the main paper) in generating high-quality pseudo-fake videos, we show some SBV samples, their blending boundaries, original landmarks, and those modified by the proposed modules in Figure \ref{fig:lmk_interpolation}. 
In the top part of the figure, we compare data generated using only CSP to data generated with both CSP and the Landmark Interpolation module. We observe that the Landmark Interpolation module ensures smooth transitions of facial landmarks between consecutive frames ($t \rightarrow t+1$). In the bottom part of the figure, we compare data generated with only CSP to data generated without any of the proposed SBV components. We observe significant variations in the manipulated facial areas when CSP is omitted. Therefore, the proposed CSP and Landmark Interpolation module effectively enhances the temporal coherence of the generated SBV.

\subsection{Impact of SBV}
To verify the advantage of using SBV for improving the generalization of detectors, we conduct several experiments using different binary classifiers trained either with SBV or with one of the four types of forgeries forming  FF++~\cite{ff++} (DF~\cite{deepfake}, F2F~\cite{face2face}, FS~\cite{faceswap}, NT~\cite{neutex}). For a fair comparison, a widely-used CNN-based Resnet3D~\cite{Resnet3D} and a Transformer-based TimeSformer~\cite{timesformer} are employed. We note that both selected models are trained from \textit{Scratch (S)} without pretrained initialization. Table.~\ref{tabl:real_vs_fake_training_full} presents the generalization performance in terms of AUC (\%) on five datasets~\cite{celeb_df, dfd, dfdcp, dfdc, wdf} respectively when trained with different manipulation methods from FF+. Notably, training with SBV significantly increases the overall generalizability capability of binary models as compared to those trained on using one specific manipulation. This indicates the importance of highly realistic, naturally consistent generated pseudo-fake videos.

\begin{table*}
\centering
\resizebox{\linewidth}{!}{
\begin{tabular}{c c ccccc H ccccccc}
\hline
\multirow{2}{*}{Method} & \multirow{2}{*}{Pretrain} & \multicolumn{5}{c}{Training set} & & \multicolumn{7}{c}{Test set AUC (\%)} \\
\cline{3-7}
\cline{9-14}
&  & Real & DF & FS & F2F & NT & & FF++ & CDF & DFD & DFDCP & DFDC & DFW & Avg. \\
\hline
\hline
ResNet3D~\cite{Resnet3D} & $\times$ & \checkmark & \checkmark & $\times$ & $\times$ & $\times$ & & \textcolor{gray}{72.5} & 58.5 & 51.3 & 53.4 & 59.4 & 65.0 & 60.0 \\
TimeSFormer~\cite{timesformer} & $\times$ & \checkmark & \checkmark & $\times$ & $\times$ & $\times$ & & \textcolor{gray}{65.4} & 59.3 & 66.1 & 53.5 & 61.4 & 57.5 & 60.5 \\
\hline

ResNet3D~\cite{Resnet3D} & $\times$ & \checkmark & $\times$ & \checkmark & $\times$ & $\times$ & & \textcolor{gray}{70.6} & 61.1 & 50.6 & 59.2 & 55.8 & 51.5 & 58.1 \\
TimeSFormer~\cite{timesformer} & $\times$ & \checkmark & $\times$ & \checkmark & $\times$ & $\times$ & & \textcolor{gray}{76.4} & 51.7 & 43.7 & 44.6 & 54.5 & 43.9 & 52.5 \\
\hline

ResNet3D~\cite{Resnet3D} & $\times$ & \checkmark & $\times$ & $\times$ & \checkmark & $\times$ & & \textcolor{gray}{78.0} & 63.8 & 54.5 & 63.4 & 55.7 & 50.1 & 60.9 \\
TimeSFormer~\cite{timesformer} & $\times$ & \checkmark & $\times$ & $\times$ & \checkmark & $\times$ & & \textcolor{gray}{81.1} & 64.4 & 60.1 & 64.5 & 52.0 & 50.5 & 62.1 \\
\hline

ResNet3D~\cite{Resnet3D} & $\times$ & \checkmark & $\times$ & $\times$ & $\times$ & \checkmark & & \textcolor{gray}{72.7} & 63.7 & 75.6 & 69.1 & 59.6 & 62.6 & 67.2 \\
TimeSFormer~\cite{timesformer} & $\times$ & \checkmark & $\times$ & $\times$ & $\times$ & \checkmark & & \textcolor{gray}{75.5} & 65.8 & 84.7 & 70.3 & 62.7 & 65.5 & 70.8 \\
\hline
\hline


ResNet3D~\cite{Resnet3D} + SBV & $\times$ & \checkmark & $\times$ & $\times$ & $\times$ & $\times$ & & \underline{90.2} & \underline{85.9} & \underline{85.0} & \underline{82.8} & \underline{66.4} & \underline{67.5} & \underline{79.6}(\textcolor{ForestGreen}{$\uparrow$12.4}) \\
TimeSFormer~\cite{timesformer} + SBV & $\times$ & \checkmark & $\times$ & $\times$ & $\times$ & $\times$ & & \textbf{94.7} & \textbf{89.5} & \textbf{95.6} & \textbf{88.6} & \textbf{72.5} & \textbf{70.9} & \textbf{85.3}(\textcolor{ForestGreen}{$\uparrow$14.5}) \\
\hline
\end{tabular}%
}
\caption{\textbf{Cross-dataset generalization}. Performance comparison in terms of AUC (\%) on multiple datasets of different binary classification models~\cite{timesformer, Resnet3D} trained using our video synthesis (SBV) and normal fake data~\cite{ff++}. All models are trained on FF++(c23)~\cite{ff++} from \textbf{\textit{Scratch (S)}} and are tested on other datasets~\cite{celeb_df, dfd, dfdcp, dfdc, wdf}. \textcolor{gray}{Gray} indicates the use of normal fake data for training. \textbf{Bold} and \underline{underline} highlight the best and the second-best performance, respectively.}
\label{tabl:real_vs_fake_training_full}
\end{table*}

\begin{figure*}
    \centering
    \includegraphics[width=\linewidth]{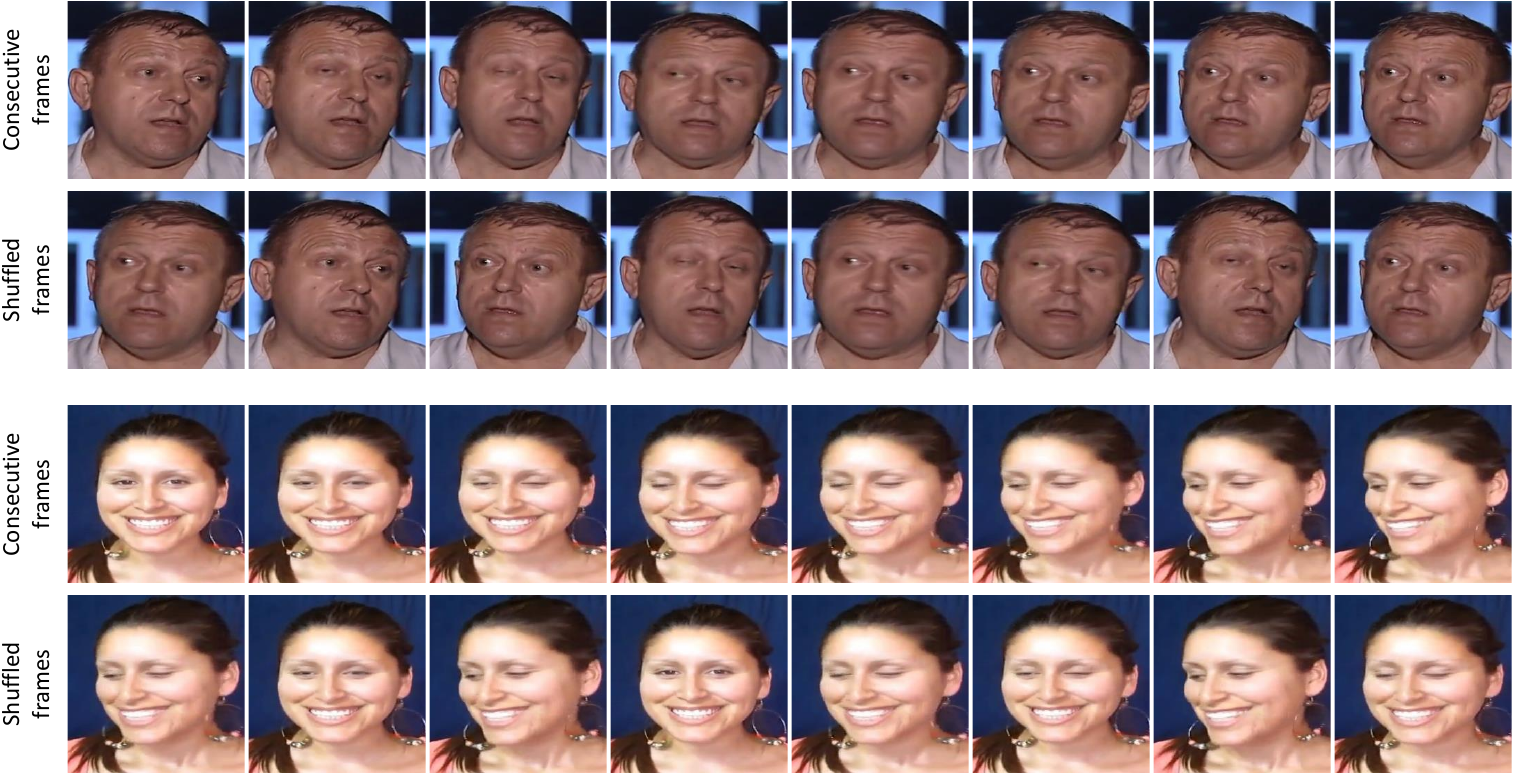}
    \caption{\textbf{Shuffled frames can produce obvious temporal inconsistencies.}}
    \label{fig:shuffled_frames}
\end{figure*}

\subsection{Robustness to Unseen Perturbations}
In the main manuscript, we report in Figure.~\textcolor{red}{5} the ``Average'' performance under different corruptions. This section complements this experiment by reporting the mean performance across different severity levels for each degradation type, as detailed in Table.~\ref{tabl:ff_noise_auc}. Except for a slight decrease in effectiveness under ``Change Saturation'' compared to LAA-Net~\cite{laa_net}, FakeSTormer is generally more robust to the unseen perturbations as compared to other augmented-based methods~\cite{laa_net, sbi, fxray, fwa}.

\begin{table*}
\centering
\scalebox{0.8}{
\begin{tabular}{c|cc|cccccc|c}
\hline
{Method} & Real & Fake & Contrast & Saturation & Gaussian Blur & Gaussian Noise & JPEG Compression & Block Wise & Avg \\ 
\hline
\hline
DSP-FWA~\cite{fwa} & $\checkmark$ & $\checkmark$ & 80.7 & 79.6 & 67.3 & 61.8 & 68.0 & 76.6 & 72.3 \\

FaceXray~\cite{fxray} & $\checkmark$ & $\checkmark$ & 88.9 & 96.0 & 70.0 & 58.0 & 62.2 & 94.7 & 78.3 \\

SBI~\cite{sbi} & $\checkmark$ & $\times$ & 92.3 & 92.0 & 72.7 & \underline{62.2} & \underline{79.1} & 92.2 & 81.7 \\

LAA-Net~\cite{laa_net} & $\checkmark$ & $\times$ & \underline{95.0} & \textbf{97.0} & \underline{73.2} & 57.5 & 75.2 & \underline{94.9} & \underline{82.1} \\
\hline
\hline

Ours & $\checkmark$ & $\times$ & \textbf{97.6} & \underline{96.3} & \textbf{81.6} & \textbf{65.4} & \textbf{92.9} & \textbf{96.8} & \textbf{88.4} \\
\hline
\end{tabular}%
}
\caption{\textbf{Robustness to unseen perturbations}. Average AUC scores (\%) across all levels for each degradation type.}
\label{tabl:ff_noise_auc}
\end{table*}

\subsection{Multi-shot Inferences}

Models can sometimes be overconfident in their predictions, which negatively impacts the generalizability aspect~\cite{mixup, label_smoothing}. To address this issue, we explore the possibility of regularizing the input during testing. Specifically, we propose multi-shot inference, leveraging Vulnerability-Driven Cutout Augmentation by utilizing the temporal head output $\Tilde{\mathbf{D}}$.
We use $\Tilde{\mathbf{D}}$ because the most significant temporal changes in vulnerable areas over time, from $t \rightarrow (t+1)$, are likely to occur at the spatial locations corresponding to the highest values of $\bar{\mathbf{B}}$ (Eq.\textcolor{red}{3} in the main manuscript).

In particular, given a test video, after the first shot of inference, the prediction map $\Tilde{\mathbf D}$ can be leveraged to generate a new masked video through the proposed Cutout augmentation for the second inference shot. Specifically, we select $\Tilde{\mathbf D}$ at $t=2$ (capturing the temporal transition from the first $\rightarrow$ the second frame) to define the set $\mathcal P$ (Section.~\textcolor{red}{3.1} of the main manuscript) for determining Cutout positions.
This iterative process can be repeated for multiple inference shots.

Table.~\ref{tabl:multi_shot_preds} presents the cross-evaluation results with five shots of inference on five unseen datasets~\cite{celeb_df, wdf, dfd, dfdcp, dfdc} using the model trained on FF++~\cite{ff++}. The results indicate a gradual improvement in generalization performance after each iteration. 
This suggests that the prediction outputs are not only interpretable but also can be used potentially to enhance the model performance.

\begin{table}
\centering
\resizebox{\linewidth}{!}{
\begin{tabular}{c cccccc}
\hline
\multirow{2}{*}{No. shots} & \multicolumn{6}{c}{Test set AUC (\%)} \\
\cline{2-6}
& CDF & DFD & DFDCP & DFDC & DFW & Avg. \\
\hline
\hline
1 & 92.35 & 98.47 & 90.02 & 74.56 & 74.19 & 85.92 \\
2 & 92.34 & \underline{98.51} & \underline{90.11} & \underline{74.60} & 74.25 & 85.96(\textcolor{ForestGreen}{$\uparrow$0.04}) \\
3 & \textbf{92.38} & \textbf{98.52} & \textbf{90.13} & \textbf{74.61} & \underline{74.28} & \underline{85.98}(\textcolor{ForestGreen}{$\uparrow$0.06}) \\
4 & \textbf{92.38} & \textbf{98.52} & \textbf{90.13} & \textbf{74.61} & \textbf{74.30} & \textbf{85.99}(\textcolor{ForestGreen}{$\uparrow$0.07}) \\
5 & \underline{92.36} & \textbf{98.52} & \textbf{90.13} & \textbf{74.61} & \textbf{74.30} & \underline{85.98}(\textcolor{ForestGreen}{$\uparrow$0.06}) \\
\hline
\end{tabular}%
}
\caption{\textbf{Multi-shot inferences.} AUC (\%) comparison of our model using different numbers of inference shots in the cross-dataset setup. The AUC slightly increases with a higher number of shots.}
\label{tabl:multi_shot_preds}
\end{table}

\subsection{Visualization of Auxiliary Branches' Outputs}

\begin{figure*}
    \centering
    \includegraphics[width=\linewidth]{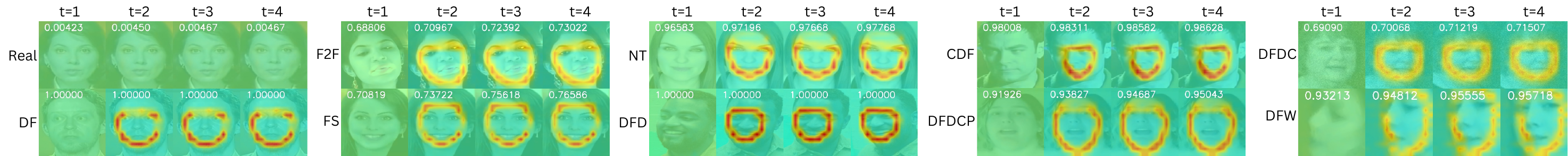}
    \caption{\textbf{Visualization of Auxiliary Branches' Outputs.} We visualize the additional auxiliary spatial and temporal branches' outputs on different unseen datasets. As shown, the number on each frame denotes the output of the spatial branch $g$, while the heatmap visualizes the output of the temporal branch $h$.}
    \label{fig:pred_output}
\end{figure*}

In addition to the probability output of the standard classification branch, FakeSTormer can provide more valuable insights from our auxiliary branches that might be conductive to prediction's post-analyses. 
Specifically, the spatial and temporal branches output the intensity of the spatial artifacts encoded in each video frame and the vulnerability change over time, respectively. The spatial branch provides frame-level scores, while the temporal branch offers more fine-grained insights. As shown in Figure~\ref{fig:pred_output}, the spatial outputs (denoted by the numbers in each frame) denote high values for fake data and low values for real data. For the temporal outputs, the heatmaps show the change in vulnerability between the instant frame $t$ and the previous one. It can be observed that it primarily focuses around the blending boundaries. We note that the change between $t-1$ and $t$ is visualized at the $t^{th}$ frame; hence, there is a \textit{blank} heatmap at the $1^{st}$ frame.

\subsection{STC~\cite{STC}: Shuffled Frames can produce obvious Temporal Inconsistencies}

We propose SBV to generate subtler artifacts for pseudo-fakes compared to the STC approach used in~\cite{STC}. We believe that STC may produce obvious (low-quality) temporal artifacts, as it shuffles frames in the temporal domain, leading to significant inconsistencies. Figure~\ref{fig:shuffled_frames} illustrates how shuffling creates noticeable discrepancies between frames. In contrast, our SBV leverages consecutive frames to produce subtler temporal artifacts while simulating these artifacts in a different manner (as detailed in Section~\textcolor{red}{3.1} of the main manuscript).

\subsection{Details on the Datasets}
\noindent\textbf{Datasets.}
For our experiments, we select datasets that haven typically used as benchmarks in previous works~\cite{altfreezing, ftcn, tall_swin, istvt, stylelatent, ucf, sfdg, ete_recons}. For both training and validation, we employ \textbf{FaceForensics++}(FF++)~\cite{ff++}, which consists of 1,000 real videos and 4,000 fake videos generated using four manipulation methods: (Deepfakes (DF)~\cite{deepfake}, FaceSwap (FS)~\cite{faceswap}, Face2Face (F2F)~\cite{face2face}, and NeuralTextures (NT)~\cite{neutex}). It can be noted that, for training, we use only the real videos and generate pseudo-fake data using our synthesized method, SBV. By default, the c23 version of FF++ is adopted, following the recent literature~\cite{ftcn, altfreezing, istvt, tall_swin, stylelatent}. 

For further validation, we also evaluate on the following datasets: (1) \textbf{Celeb-DFv2} (CDF)~\cite{celeb_df}, a well-known benchmark with high-quality deepfakes; (2) \textbf{DeepfakeDetection} (DFD)~\cite{dfd}, which includes 3,000 forged videos featuring 28 actors in various scenes; (3) \textbf{Deepfake Detection Challenge Preview} (DFDCP)~\cite{dfdcp} and (4) \textbf{Deepfake Detection Challenge} (DFDC)~\cite{dfdc}, a large-scale dataset containing numerous distorted videos with issues such as compression and noise; (5) \textbf{WildDeepfake} (DFW)~\cite{wdf}, a dataset fully sourced from the internet, without prior knowledge of manipulation methods; (6) \textbf{DiffSwap} generated in the similar protocol as in LFGDIN~\cite{LFGDIN} by using a recent diffusion-based swapping method~\cite{DiffSwap} on $250$ real videos selected from CDF~\cite{celeb_df}; and (7) \textbf{DF40}~\cite{DF40}, a highly diverse and large-scale dataset comprising $40$ distinct deepfake techniques, enables more comprehensive evaluations for the next generation of deepfake detection.

\noindent\textbf{Data Pre-processing.} Following the splitting convention~\cite{ff++}, we extract $256$, $32$, and $32$ consecutive frames for training, validation, and testing, respectively. Facial regions are cropped using Face-RetinaNet~\cite{retina_face}. These bounding boxes are conservatively enlarged by a factor of $1.25$ around the center of the face and then resized to a fixed resolution of $224 \times 224$. Additionally, we store $81$ facial landmarks for each frame, extracted using Dlib~\cite{dlib}. Finally, the preserved landmark keypoints are utilized to dynamically generate pseudo-fakes during each training iteration.

\subsection{Revisited TimeSformer: Implementation Details}
We choose TimeSformer~\cite{timesformer} as our feature extractor given its ability to effectively capture separate long-range temporal information and spatial features.
First, given a video $\mathbf X \in \mathbb R^{C \times T \times H \times W}$, its frames in each time step are split into $N$ number of non-overlapping patches of size $P \times P$, i.e., $N = \frac{H \times W}{P^2}$. Each patch is flatten as $\mathbf x_{(t,p)} \in \mathbb R^{C.P^2}$, and is then linearly mapped into $D$-dimensional embedding vector $\mathbf z_{(t,p)}^0 \in \mathbb R^{D}$ by means of a learnable matrix $E \in \mathbb R^{D \times C.P^2}$ where $t = [[1,T]]$ indexes temporal positions, and $p = [[1,N]]$ indexes spatial positions. The process results in an input patch embedding matrix $\mathbf Z^0 \in \mathbb R^{T \times N \times D}$.

In TimeSformer, a global class token $\mathbf z_{cls}$ attends to all patches and then is used for classification. This mechanism implicitly captures mixed spatial-temporal features at the same time, which might lead to overfitting on a specific type of domain artifacts~\cite{ftcn, altfreezing}. We revisit it slightly in order to decouple the spatial and temporal information by considering two sorts of additional tokens (one spatial and one temporal).

For that purpose, we attach in each dimension of $\mathbf Z^0$, a spatial token $\mathbf z_s^0 \in \mathbb R^D$ and a temporal token $\mathbf z_t^0 \in \mathbb R^D$, respectively. These tokens will independently interact only with patch embeddings belonging to their dimension axis by leveraging the decomposed SA~\cite{timesformer}. This mechanism not only facilitates the disentanglement learning process of spatio-temporal features but is also beneficial to optimize the computational complexity of $\mathcal O(T^2 + N^2)$ as compared to $\mathcal O(T^2 \cdot N^2)$ in vanilla SA. 
Those tokens will be then fed into $L$ ($L=12$ as default) transformer encoder blocks in which each block contains a multi-head temporal SA (TSA), a multi-head spatial SA (SSA), LayerNorm (LN), and a multi-layer perception (MLP). Note that, for the sake of matrix compatibility, a placeholder embedding $\mathbf z_{(0,0)}^0$ is attached.
Formally, the feature extraction process can be summarized as follows,
\begin{equation}
    [\mathbf Z^L, \mathbf z^L_s, \mathbf z^L_t ] = \Phi( \mathbf X),
\end{equation}

\noindent where $\mathbf Z^L$ is the final patch embedding matrix, $\mathbf z^L_s$ the resulting set of spatial tokens, and $\mathbf z^L_t$ the resulting set of temporal tokens that will be respectively sent to the temporal head $h$, the spatial head $g$, and the classification head $f$. Our overall framework is illustrated in Figure.~\textcolor{red}{2} of the main paper.

\end{document}